\documentclass{article}
\usepackage[utf8]{inputenc}

\usepackage{amsmath,amsthm} 
\usepackage{amssymb,mathrsfs} 
\usepackage{a4wide} 
\usepackage{graphicx} 
\usepackage{color} 
\usepackage{subcaption}
\usepackage{enumerate}  
\usepackage{fullpage}
\usepackage{algorithm,algorithmic}
\usepackage{bbm}
\usepackage[colorlinks,citecolor=blue,urlcolor=blue]{hyperref}
\usepackage{cleveref}
\usepackage{todonotes}

\DeclareMathOperator*{\argmax}{arg\,max}
\DeclareMathOperator*{\argmin}{arg\,min}
\newtheorem{theorem}{Theorem}

\newtheorem{remark}[theorem]{Remark}

\DeclareMathAlphabet{\mathpzc}{OT1}{pzc}{m}{it}

\newcommand{\eps}{\varepsilon}

\newcommand{\RR}{\mathbb{R}}
\newcommand{\R}{\mathbb{R}}

\newcommand{\dt}{{\Delta t}}
\renewcommand{\leq}{\leqslant}
\renewcommand{\geq}{\geqslant}

\begin{document}

  
\title{Generative methods for sampling transition paths in molecular dynamics}

 \author{T. Leli{\`e}vre$^{1,2}$, G. Robin$^{3,4}$, I. Sekkat$^{1}$, G. Stoltz$^{1,2}$ and G. Victorino Cardoso$^{4,5}$\\
  {\small $^{1}$ CERMICS, Ecole des Ponts, Marne-la-Vall\'ee, France} \\
  {\small $^{2}$ MATHERIALS team-project, Inria Paris, France} \\
  {\small $^3$ CNRS \& Universit\'e d'Evry, France} \\
  {\small $^{4}$ CMAP, Ecole Polytechnique, Palaiseau, France} \\
  {\small $^{5}$ LIRYC, Université de Bordeaux, Bordeaux, France} \\
}
\maketitle

\abstract{
  Molecular systems often remain trapped for long times around some local minimum of the potential energy function, before switching to another one -- a behavior known as metastability. Simulating transition paths linking one metastable state to another one is difficult by direct numerical methods. In view of the promises of machine learning techniques, we explore in this work two approaches to more efficiently generate transition paths: sampling methods based on generative models such as variational autoencoders, and importance sampling methods based on reinforcement learning.
}

\section{Introduction}\label{sec:intro}

Molecular dynamics aims at simulating the physical movement of atoms in order to sample the Boltzmann--Gibbs probability measure and the associated trajectories, and to compute macroscopic properties using Monte Carlo estimates~\cite{frenkel-smit-02, AT17}. One of the main difficulties when performing these numerical simulations is metastability: the system tends to stay trapped in some regions of the phase space, typically in the vicinity of local maxima of the target probability measure. In this context, transitions from one metastable state to another one are of particular interest in complex systems, as they characterize for example crystallisation or enzymatic reactions. These reactions happen on a long time scale compared to the molecular timescale, so that the simulation of realistic rare events is computationally difficult.

On the one hand, many efforts have been devoted to the development of rare events sampling methods in molecular dynamics. The goal of these methods is to characterize transition paths and to compute associated transition rates and mean transition times; see for instance~\cite{e16010350} for a review. The most notable methods can be classified in two groups: 
\begin{enumerate}[(i)]
\item importance sampling techniques, where the dynamics is biased (by modifying the potential for instance) to reduce the variance of Monte Carlo estimators when computing expectations, see for instance~\cite{Fleming77,Bucklew04} for more details, and also~\cite[Section~6.2]{lelievre_stoltz_2016}. It is possible to use adaptive importance sampling strategies to choose the importance function, see~\cite[Chapter~5]{lelievre-rousset-stoltz-book-10}. Another viewpoint is offered by the framework of stochastic control, as in~\cite{e16010350} where the modification in the drift of the dynamics is determined by the solution of an optimal control problem. 
\item splitting methods, where the idea is to decompose the rare event to sample as a succession of moderately rare events. In the context of trajectories relating two local maxima of the target probability measure, this can be done in an adaptive manner using the so-called Adaptive Multilevel Splitting algorithm, where an ensemble of trajectories are concurrently evolved, removing the ones that lag behind in terms of progress towards the target state, and replicating the ones exploring more successfully the path towards the target state; see~\cite{doi:10.1080/07362990601139628, aristoff2015adaptive, cerou2011multiple, cerou2019adaptive}.
\end{enumerate}

On the other hand, generative models aim at generating samples whose distribution approximates some unknown target distribution. They have attracted a lot of attention lately due to their wide range of applications, such as text translation, out-of-distribution detection, generation of new human poses, etc. Currently, popular generative models are Generative Adversarial Networks (GANs)~\cite{Goodfellow2014GANs}, Variational AutoEncoders (VAEs)~\cite{Kingma2013VB, Rezende2014}, as well as Energy-Based Models and their extensions; see~\cite{bond2021deep} for a review of the most important models. Diffusion models also recently appeared and attracted a lot of attention~\cite{DiffusionModels}. Generative models have also been used in the context of rare event sampling. For instance, GANs can be used to generate data from extreme tails of (heavy tailed) distributions, as discussed in~\cite{allouche2021tail} and references therein. Generative models also offer the perspective to detect anomalies which can be considered as rare events~\cite{an2015variational, dionelis2020tail}, or maybe even generate anomalous states by sampling from outlier regions in the latent space~\cite{anogen}. In the context of molecular dynamics, machine learning techniques have been used to study transition pathways~\cite{varolgunecs2020interpretable, rotskoff2020learning, zeng2021note}. In~\cite{rotskoff2020learning}, the authors suggest to use a neural network to approximate the committor function giving the probability of reaching a metastable state before another one, importance sampling techniques being used to reduce the statistical error in these computations. VAEs have been used in~\cite{varolgunecs2020interpretable} to find collective variables by considering mixtures of Gaussian priors in the latent space to encode the trajectories.

The goal of this work is to explore some machine learning techniques to efficiently generate transition paths in molecular dynamics. We first tried a data-driven generative method: from a given data set of transition paths, we learn to generate new ones using variational autoencoders. Using VAEs naively, the temporal aspect of the trajectories is not encoded in the latent variables, which produces unconvincing results when generating new trajectories.  We tested two techniques to learn the temporal aspect on the latent space, namely vector quantized variational autoencoders~\cite{oord2018neural} and variational recurrent neural network~\cite{chung2016recurrent} but these approaches were not successful. We therefore turned to a data free approach, relying on reinforcement learning algorithms to construct trajectories following the dynamics introduced in~\eqref{eq:Langevin-discrete}, while guiding it to transition from one well to another one. Reinforcement learning is more convenient than generative approaches learning from a dataset when the construction of the data set is computationally challenging. 

This work is organized as follows. We introduce the main settings of the molecular dynamics problem we tackle in Section~\ref{sec:sampling_pres}. In Section~\ref{sec:vae}, we briefly present variational autoencoders, and the methods used to learn the temporal aspect on the latent space. Section~\ref{sec:RL} is dedicated to results obtained with reinforcement learning. 

\section{Sampling transition paths of metastable processes}\label{sec:sampling_pres}
We present in this section the main settings of the problem we tackle.
\paragraph{Sampling from the Boltzmann--Gibbs distribution.}  Let us consider a diffusion process~$(q_t)_{t \geq0}$ with values in~$ \mathcal{D}  = \RR^d$,  whose drift derives from a potential~$V:\mathcal{D} \rightarrow \RR$.  We typically consider the case when the potential~$V$ has many local minima. We want to sample from the Boltzmann-Gibbs distribution given by~$\mu(dq) =Z^{-1} \mathrm{e}^{-\beta V(q)}\, dq$. In this case, one of the main issues when sampling trajectories is metastability: the system remains trapped for a long time around some local minimum of~$V$ before jumping to another local minimum. Our goal is to simulate \textit{transition paths}, that we define in this work as trajectories which, from a fixed initial condition $q_0$ located in an initial potential well $A$, reach a pre-specified set $B\subset \RR^d$ before time $T\geq 0$. Typically, $B$ corresponds to another well in the energy landscape.

\paragraph{Overdamped Langevin dynamics.} The evolution of molecular systems can be modelled by Langevin dynamics, which are stochastic perturbations of the Hamiltonian dynamics. For simplicity in this work, we consider that the system evolves according to the overdamped Langevin diffusion
\begin{equation}
    \label{eq:O_Langevin}
    dQ_t = -\nabla V(Q_t)\,dt + \sqrt{\frac{2}{\beta}}\,dW_t,
\end{equation}
where $(W_t)_{t \geq 0}$ is a standard $d$-dimensional Wiener process. The dynamics~\eqref{eq:O_Langevin} admits the Boltzmann--Gibbs distribution as a unique invariant probability measure (see for instance~\cite{MR885138}). 
In practice, we use a Euler--Maruyama discretization with a time step $\dt > 0$ to approximate the exact solution of the stochastic differential equation~\eqref{eq:O_Langevin}. We obtain the following discrete-time process:
\begin{equation}
    \label{eq:Langevin-discrete}
    q_{k+1} = q_k -\nabla V(q_k)\Delta t + \sqrt{\frac{2\Delta t}{\beta}}G_k,
\end{equation}
where~$G_k \sim \mathcal{N}(0,\mathrm{I}_d)$ for all~$k\geq 0$ are independent Gaussian random variables. We assume that the drift of the dynamics is globally Lipschitz or that Lyapunov conditions are satisfied, so that the Markov chain corresponding to the time discretization~\eqref{eq:Langevin-discrete} admits a unique invariant probability measure, denoted by~$\mu_{\Delta t}$; see~\cite{MR1931266}. It is well known that the Euler--Maruyama discretization~\eqref{eq:Langevin-discrete} is consistent (weakly and strongly) of order~$1$, and that~$\mu_{\Delta t}$ agrees with~$\mu$ up to errors of order~$\Delta t$ (see for instance~\cite{doi:10.1080/17442509008833606} and~\cite[Theorem~7.3]{MR1931266} for the latter point).

\begin{figure}
\centering
    \centering
    \includegraphics[scale=0.6]{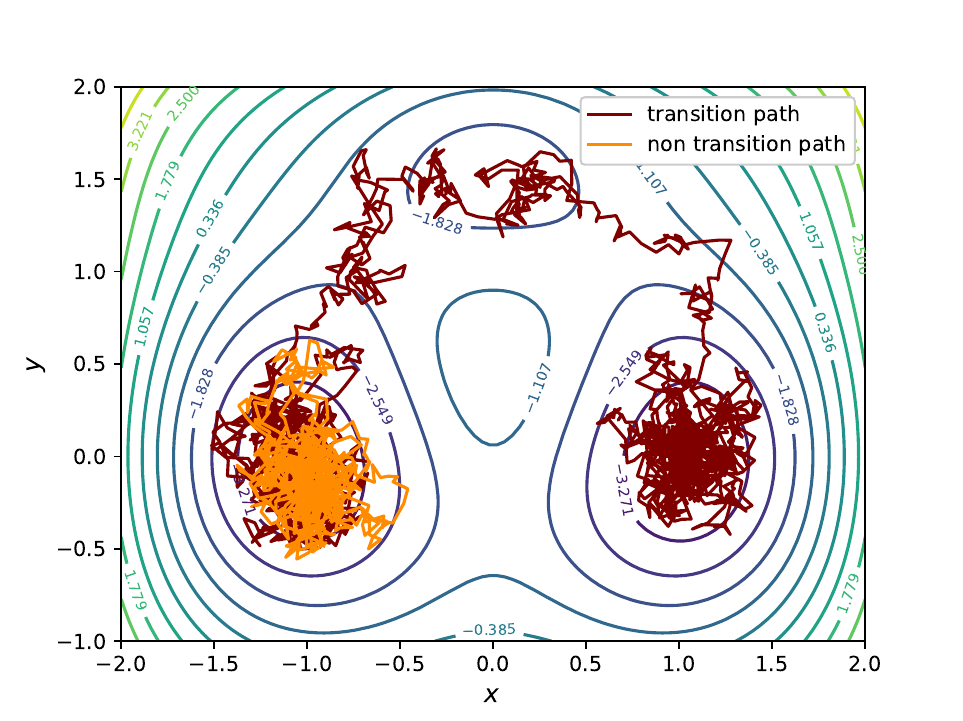}
    \caption{Transition path and non transition path in the 2-dimensional potential given by~\eqref{eq:potentiel_2d}.}
    \label{fig:2Dtraj}
\end{figure}

\paragraph{Two--dimensional numerical example.} To illustrate the metastability issue, we present a simple two dimensional example, which will be the running numerical example of this paper. We assume that $\mathcal{D} = \mathbb{R}^2$, and consider the following potential for $q=(x, y)$ (already used in~\cite{park2003reaction, doi:10.1137/070699500}):
\begin{equation}
\begin{aligned}
V(q)  = &\  3 \exp\left(-x^2 - \left(y-\frac13\right)^2\right) - 3 \exp\left(-x^2 - \left(y-\frac53\right)^2\right) -5 \exp\left(-\left(x-1\right)^2 - y^2\right) \\
& - 5 \exp\left(-\left(x+1\right)^2 - y^2\right) + 0.2 x^4 + 0.2 \left(y-\frac13\right)^4.
\label{eq:potentiel_2d}
\end{aligned}
\end{equation}
We plot in Figure~\ref{fig:2Dtraj} two trajectories generated using the discretization~\eqref{eq:Langevin-discrete}, with $\dt = 5 \times 10^{-3}$, $\beta = 3.5$ and a final time $T = 10$. The first trajectory, displayed in orange, remains trapped in the first well $A$ (located around $(-1, 0)$), whereas the second trajectory, displayed in red, jumps to the second well $B$ by going through the local minimum of~$V$ on the top. An alternative path for the particle to go from the first to the second well passes through the bottom.
We performed direct numerical simulations of trajectories of length~$T=10$ initialized from~$(-1.05,-0.04)$ to confirm that sampling paths transitioning from~$A$ to~$B$ is rare, with a probability of the order of~0.01. Directly integrating~\eqref{eq:Langevin-discrete} to explore the configurational space is therefore inefficient in this case. The issue becomes even more acute in higher dimensions, especially when the potential $V$ has many local minima. 

\section{Generating transition paths with Variational AutoEncoders}
\label{sec:vae}

The purpose of this section is to use a data set of transition paths to generate new ones using variational autoencoders (VAEs). We first present VAEs in Section~\ref{sec:pres_vae}. We briefly recall in Section~\ref{sec:CNN} the convolutional layers which are the building blocks of the various architectures used in this section. The construction of the data set used to train the various models is discussed in Section~\ref{sec:data}. We then present in Section~\ref{sec:2d_vae} the 2-dimensional VAE we consider, alongside with the numerical results for this model. Finally, in Section~\ref{sec:64d_vae} we describe some methods to incorporate the temporal aspect of the data in the latent space, first with a ``naive" VAE, and then using  vector quantized VAEs (VQ-VAEs) and variational recurrent neural networks. 

\subsection{Presentation of Variational AutoEncoders}
\label{sec:pres_vae}

Variational autoencoders, as introduced by~\cite{Kingma2013VB,Rezende2014} and reviewed in~\cite{KW19}, are a class of generative models based on latent variables. For VAEs, these latent variables correspond to values of the input encoded by the neural network used in practice. Assume that the data consists of $n$ observed variables, which we denote by $\mathbf{q} = (q^{1},\ldots, q^{n})\in \mathcal{D}^{n}$, distributed according to some probability measure $p(\cdot)$. In our context, each element~$q^i$ is a time-discrete trajectory, \emph{i.e.} a time ordered sequence of configurations of the system (see Section~\ref{sec:data} for a more precise description). The dimension of the data is, in this case, the product of the number of configurations appearing in the time discrete trajectory multiplied by the physical dimension of each configuration.
 
The aim is to approximate the distribution of the data by a parametric distribution $p_\theta \approx p$. Instead of considering simple but limited parametric distributions, generative models assume that there exist latent, unobserved variables, which we denote by $\mathbf{z} = (z^{1},\ldots, z^{n})\in (\mathbb{R}^{\ell})^{n}$, where $\ell$, the dimension of the latent variables (called the intrinsic dimension), is generally smaller than the dimension of the data. In this context, the likelihood is given by 
 \begin{equation}
 p_\theta(q) = \int_{\R^\ell} p_\theta (q, z) \, dz = \int_{\R^\ell} p_\theta (q|z) p_\theta(z) \, dz. 
 \label{eq:max_likelihood}
 \end{equation}
The aim is to maximize the likelihood of $(q^1, q^2, ..., q^n)$ with respect to $\theta$. Note that, with some abuse of notation, the joint distributions of~$z$ and~$q$, the marginal distributions in~$q$ and~$z$ and the prior distribution on~$z$ are all denoted by~$p_\theta$. The joint distribution of $(\mathbf{q},\mathbf{z})$ is defined through a parametric model with unknown parameter $\theta$ as
 \begin{equation}
 \label{eq:dllvm}
 \begin{split}
 z^{i} & \sim p_{\theta}(z), \\
 q^{i}|z^{i} & \sim p_{\theta}(q|z^{i}) .
 \end{split}
 \end{equation} 

\paragraph{Evidence lower bound.} In the setting considered here, computing the likelihood~\eqref{eq:max_likelihood}, as well as the conditional probability $p_\theta(z|q)$, is intractable. In view of Bayes' relation, the likelihood~$p_\theta(q)$ and the conditional likelihood~$p_\theta(z|q )$ are related as
\[ p_\theta(q) = \frac{p_\theta(q, z)}{ p_\theta(z|q)} =  \frac{p_\theta(q|z) p_\theta(z)}{ p_\theta(z|q)}.
\]
From a Bayesian perspective, one of the aims of VAEs is the inference of the posterior distribution of the latent variables, $p_{\theta}(z|q)$. To do so, VAEs rely on \textit{variational inference} (see, e.g., \cite{Blei2017VarInference}), which can be seen as an alternative to Markov Chain Monte Carlo sampling in complex Bayesian models where $p_{\theta}(z|q)$ is intractable. The idea of variational inference is to posit a family of probability distributions $\Pi$, and to approximate the posterior $p_{\theta}(z|q)$ by a distribution in the family~$\Pi$ of distributions in $z$ indexed by $q$ which minimizes the Kullback--Leibler divergence
\begin{equation}
\label{eq:var-inference}
\pi^{\star} = \mathop{\mathrm{argmin}}_{\pi\in\Pi}\mathsf{KL}\left(\pi(z|q)\middle||p_{\theta}(z|q)\right), \qquad\mathsf{KL}(\pi(z|q)||p_{\theta}(z|q))   = \int_{\R^\ell} \pi(z|q) \log\left( \frac{\pi(z|q)}{p_{\theta}(z|q)} \right) \,dz  .
\end{equation}
A simple computation shows that
\begin{equation}
    \label{eq:var-objective}
    \begin{split}
        \mathsf{KL}(\pi(z|q)||p_{\theta}(z|q)) & = \mathbb{E}_{\pi(\cdot|q)}[\log \pi(z|q)] - \mathbb{E}_{\pi(\cdot|q)}[\log p_{\theta}(z|q)]\\
        & = \mathbb{E}_{\pi(\cdot|q)}[\log \pi(z|q)] - \mathbb{E}_{\pi(\cdot|q)}[\log p_{\theta}(z,q)]+\log p_{\theta}(q).
    \end{split}
\end{equation}
Reordering the terms in \eqref{eq:var-objective}, we obtain:
\begin{equation}
  \label{eq:ELBO}
  \log p_{\theta}(q)-\mathsf{KL}(\pi(z|q)||p_{\theta}(z|q)) = \underbrace{\mathbb{E}_{\pi(\cdot|q)}[\log p_{\theta}(z,q)] - \mathbb{E}_{\pi(\cdot|q)}[\log \pi(z|q)]}_{\mathsf{ELBO}}.
\end{equation}
The term on the right hand side of the previous equality is called the \textit{Evidence Lower Bound} (ELBO). Maximizing this quantity achieves two aims:
\begin{itemize}
\item since $\mathsf{KL}(\pi(z|q)||p_{\theta}(z|q))\geq 0$, the ELBO is a lower bound on the marginal log-likelihood $\log p_{\theta}(q)$. Thus, maximizing the ELBO with respect to $\theta$ provides a proxy for the maximum likelihood estimate of parameter $\theta$. Note that this is an important goal, since approximating~$\theta$ yields an approximation to the distribution of the observed data~$\mathbf{q}$, and thus allows to generate new samples.
\item 
  Maximizing the ELBO also leads to decreasing~$\mathsf{KL}(\pi(z|q)||p_{\theta}(z|q)) \geq 0$. In view of~\eqref{eq:var-inference}, this provides an approximation of the intractable posterior distribution~$p_\theta(z|q)$, called ``encoder'' or ``recognition model''. The better this approximation is, the tighter the ELBO lower bound is. 
\end{itemize}

\paragraph{Variational autoencoders.}
The idea of variational inference is to choose a class of distributions~$\Pi$ which is consistent with our intuition of the problem, and yields a tractable optimization problem. Variational Autoencoders rely on a different class of distributions~$\Pi$, defined as follows:
\begin{equation}
\label{eq:post-model}
\pi(z|q) = \Psi(z|g_{\phi}(q)).
\end{equation}
In \eqref{eq:post-model}, $\{\Psi(.|\gamma), \gamma \in \mathscr{G}\}$ is a parametric family of densities, and $g_{\phi}: \mathcal{D}\rightarrow \mathscr{G}$ is a differentiable function. For instance, $\Psi(z|\mu, \Sigma)$ can be chosen as the multivariate Gaussian distribution with mean $\mu$ and covariance matrix $\Sigma$. In this case, $g_{\phi}(q) = (\mu_{\phi}(q), \Sigma_{\phi}(q))$, where $\mu_{\phi}$ and $\Sigma_{\phi}$ are parametrized by a neural network with weights $\phi$. The function $g_{\phi}$ is referred to as the \textit{encoder}: it allows to construct the distribution of the latent variables, given the observations. 

\paragraph{Autoencoding variational bound algorithm.} Assembling the concepts of the previous paragraphs, VAEs aim at solving the following optimization problem:
\begin{equation}
\label{eq:VAE-2}
\mathop{\mathrm{argmax}}_{\theta, \phi} \sum_{i=1}^{n} \mathcal{L}(\theta, \phi; q^{i}), 
\end{equation}
where
\[
\mathcal{L}(\theta, \phi; q^{i}):=\mathbb{E}_{z\sim \Psi(.|g_{\phi}(q^{i}))}\left[\log p_{\theta}\left(q^{i}\middle|z\right) \right]- \mathbb{E}_{z\sim \Psi(.|g_{\phi}(q^{i}))}\left[ \log \Psi\left(z\middle|g_{\phi}(q^{i})\right) - \log(p_\theta(z))\right],
\]
where~$(q^1, ..., q^n)$ is the given data set of observed quantities. Problem \eqref{eq:VAE-2} is solved using a stochastic gradient descent (SGD) algorithm. While computing the gradient with respect to $\theta$ is straightforward, using a Monte Carlo estimator to compute the gradient of $\mathcal{L}(\theta, \phi; q^{i})$  with respect to $\phi$ however leads to a large variance~\cite{paisley2012variational, Kingma2013VB}. It is suggested in~\cite{Kingma2013VB} to use the so-called reparametrization trick to obtain expressions of gradients both with respect to $\phi$ and $\theta$. The method goes as follows. Considering a diagonal covariance matrix~$\Sigma_\phi = \mathrm{diag}(\sigma^2_\phi)$ with $\sigma_\phi \in \R^\ell$, and since $\{\Psi(.|\gamma), \gamma \in \mathscr{G}\}$ corresponds to a family of multivariate Gaussian distributions $\Psi(q|\mu_\phi, \Sigma_\phi)$ with mean~$\mu_\phi$, the random variable $z\sim\Psi(\cdot|g_{\phi}(q^{i}))$ can be reparametrized as
\begin{equation}
\label{eq:reparam}
z = \mu_{\phi}(q^{i}) + \mathrm{diag}(\sigma_{\phi}(q^{i}))\eps, \qquad \eps\sim \mathcal{N}(0, \text{I}_\ell).
\end{equation}
Note that similar reparametrizations can be considered whenever $\Psi$ is a ``location-scale" family of distribution (Laplace, Student, etc.). Alternatively, if $\Psi$ has a tractable CDF, one can reparametrize $z$ with a uniform random variable $\eps\sim\mathcal{U}([0,1])$. With the choice~\eqref{eq:reparam},
\begin{equation}
\label{eq:GaussianVAE}
\begin{aligned}
\mathcal{L}(\theta, \phi; q^{i})& =\mathbb{E}_{\eps\sim \mathcal{N}(0,1)}[\log p_{\theta}(q|\mu_{\phi}(q^{i}) + \sigma_{\phi}(q^{i})\eps)]\\
  & \ \ - \mathbb{E}_{\eps\sim \mathcal{N}(0,1)}[\log \Psi(\mu_{\phi}(q^{i}) + \sigma_{\phi}(q^{i})\eps|g_{\phi}(q^{i})) - \log p_\theta(\mu_{\phi}(q^{i}) + \sigma_{\phi}(q^{i})\eps)].
\end{aligned}
\end{equation}
 To make the loss function~\eqref{eq:GaussianVAE} even more explicit, we consider that the prior $p_\theta(z)$ is a centered reduced Gaussian distribution. In this case, the term on the second line, which corresponds to the Kullback--Leibler divergence of two Gaussian distributions, can be analytically computed. The term on the first line can be approximated by a Monte Carlo discretization using $(\eps^1,\ldots, \eps^{L})$ independent $\ell$-dimensional standard normal Gaussian vectors. This amounts to considering the following estimator $\widehat{\mathcal{L}}(\theta, \phi; q^{i}) $ of $\mathcal{L}(\theta, \phi; q^{i})$: 
\begin{equation}
\label{eq:MC-approx}
\widehat{\mathcal{L}}(\theta, \phi; q^{i}) =\frac{1}{L}\sum_{j=1}^L\log p_{\theta}(q^{i}| \mu_{\phi}(q^{i}) + \sigma_{\phi}(q^{i})\eps^{j}) +  \sum_{k = 1}^\ell\left[1+\log(\sigma_{\phi}^2(q^{i})_k)-\mu_{\phi}^2(q^{i})_k - \sigma^2_{\phi}(q^{i})_k\right],
\end{equation}
where we denote by~$\mu_\phi(q)_k$ and~$\sigma_\phi(q)_k$ the components of the vectors~$\mu_\phi(q)$ and~$\sigma_\phi(q)$. In practice, we consider $L=1$ as suggested in~\cite{Kingma2013VB}. We also assume that $p_\theta (q| z) = \Phi(q|f_\theta(z))$, where $\{\Phi(\cdot|\eta), \eta \in \mathscr{H}\}$ is a family of densities parametrized by $\eta \in \mathscr{H}$. For instance, one can consider for $\Phi(q|\mu, \Sigma)$ the multivariate Gaussian distribution with mean $\mu$ and covariance matrix $\Sigma$, and $f_{\theta}(z) = (\mu_{\theta}(z), \Sigma_{\theta}(z))$ where $\Sigma_\theta(z)  = \sigma^2_\theta(z)\mathrm{I}_d$. This particular case amounts to considering $p_\theta$ as an infinite mixture of Gaussians. The  function $f_{\theta}: \mathbb{R}^\ell \rightarrow \mathscr{H}$ is called the \textit{decoder}: it allows to reconstruct the distribution of the data given the latent variables. In general, one assumes $f_{\theta}$ to be parametrized by a neural network with weights $\theta$. Note that in this specific Gaussian setting,
\begin{equation}
\label{eq:log-marginal-MC}
\log p_{\theta}(q| z) = -\frac{d}{2}\log(2\pi) - d\log(\sigma_{\theta}(z)) - \frac{\|q-\mu_{\theta}(z)\|^2}{2\sigma_{\theta}^2(z)},
\end{equation}
where $z$ is given by~\eqref{eq:reparam}. In fact, for the VAEs we used, we considered $\sigma_{\theta}$ as a constant and fixed it to $\sigma_{\theta} = 1.2 \times 10^{-2}$.
As a result, the terms in \eqref{eq:MC-approx} and \eqref{eq:log-marginal-MC} are differentiable with respect to $\phi$ and $\theta$, and the gradients can be computed using backpropagation. 

To summarize, we use a neural network as the encoder, which takes as input a trajectory $q^i$, and gives as output $\mu_{\phi}(q^{i}), \sigma_{\phi}(q^{i})$. We use the reparametrization trick to compute the latent variable $z^i$, which is given as input to the decoder (which is itself a neural network). The output of the decoder is the trajectory~$\mu_\theta(z^i)$. The loss function in~\eqref{eq:VAE-2} (using in particular~\eqref{eq:log-marginal-MC}) can then be minimized with respect to $\theta, \phi$ with gradients computed using backpropagation. Gradients are approximated in practice by minibatching. The building block of the encoder and decoder we use are convolutional neural networks presented in the next section.

\subsection{Convolutional neural networks}
\label{sec:CNN}

Convolutional neural networks~\cite{lecun1989backpropagation, lecun1998gradient} are generally used for data invariant under translation and scaling. They are composed of hidden stacked layers: convolutional layers, pooling layers, normalization layers, to which one generally adds a fully connected layer at the end. We describe more precisely convolutional layers in this section, since they are key building blocks used in the various architectures we consider (alongside some fully connected layers and batch normalization layers). The input, which corresponds to the first layer, is a trajectory~$q$.

A convolutional layer transforms an input time series $(X_t)_{1 \leq t\leq T_{\mathrm{in}}} \in \R^{T_{\mathrm{in}}\times m_{\mathrm{in} }}$ into a time series $Y \in \R^{T_{\mathrm{out}} \times m_{\mathrm{out} }}$. The parameters of a convolutional layer are the elements of the $m_{\mathrm{out} }$ convolution kernels of size $k \times m_{\mathrm{in} }$, denoted by $W \in \mathbb{R}^{k \times m_\mathrm{in} \times m_\mathrm{out} }$, with~$k<T_{\rm in}$. More precisely, each convolution kernel $W^{j} \in \R^{k \times m_{\mathrm{in} }}$ for $j \in \{1, ..., m_{\mathrm{out}} \}$ performs Frobenius products (denoted by $:$) with parts of the input time series. The kernel $W^{j}$ convolves through the input to form a new time series, which is then passed to an activation function. The stride $s$ controls the shift of the kernel through the input time series. More precisely, $Y^{j}_t$ is defined for $1\leq t \leq T_\mathrm{out}$ and $1\leq j \leq m_\mathrm{out}$ as
\begin{equation*}\label{eq:CNN_formula}
Y^{j}_t = f\left(W^{j} : X_{[(t-1)s+1; (t-1)s + k]}\right), 
\end{equation*}
where~$f$ is an activation function. For all the models considered in this work, we use the ReLU function~$f(z) = \max (0,z)$.

There are several important things to notice in~\eqref{eq:CNN_formula}. The first one is that the output time series $(Y_t)_{1\leq t \leq T_\mathrm{out}}$ does not have the same dimension as the input time series. Its length $T_{\mathrm{out}}$ is defined by the following formula:
\begin{equation}
T_{\mathrm{out}} = \frac{T_{\rm in} - k}{s} + 1.
\end{equation}
If we want it to have the same size as the input, we should set $s=1$ and use some padding to compensate for the missing entries. The learnable parameter is $W=(W^1,\dots,W^{m_{\mathrm{out} }})$, and the other parameters are defined by the user (including $k$ and $m_{\mathrm{out} }$).  The integer~$m_{\mathrm{in} }$ is the number of channels of the input and $m_{\mathrm{out} }$ the number of channels of the ouput. In our model, we use a convolutional neural network obtained by stacking together convolutional layers. In this context, the receptive field is defined as the number of elements of the input time series contributing to the value of one element of the output. It is traditionally considered that a good network is one that has a receptive field of the size of the largest characteristic scale (the characteristic time of the transition from $A$ to $B$ here), while also reducing the temporal complexity. A trade off from an optimization perspective is to have $T_{\mathrm{out}} \ll T$ and $m_{\mathrm{out} } \gg d$. 

\subsection{Data set for training}\label{sec:data}
To train VAEs, one needs a data base of transition paths. We use the system described in Section~\ref{sec:pres_vae} for which we generate $12,968$ trajectories. We start from a fixed initial condition $q_0 = ( -1.05, -0.04)$, and integrate the dynamics using~\eqref{eq:Langevin-discrete} with the potential~\eqref{eq:potentiel_2d} to construct a discrete trajectory $(q_0,q_1,\dots,q_T)$, where $T=1984$. We finally classify the trajectories as transitions through the bottom, transitions through the top and absence of transition. A trajectory is considered a transition if there exists $k \in \{1, ..., T\}$ such that $x_k>0$ (where we recall~$q_k=(x_k,y_k)$). If~$y_k$ is greater than~$0.7$, the trajectory is classified as transitioning through the top, otherwise it is classified as transitioning through the bottom. The data set used to train all the models in this section is composed of $1,743$ transition paths ($510$ with a transition through the top and $1,233$ through the bottom) and $11,225$ non transition paths. We use $80\%$ of the data to train the models, whereas $20\%$ of the data is used for testing. We also tried to train the models without the non transition paths from the data set, but the results were very similar. Let us emphasize here that constructing a data set containing transition paths can be computationally hard and possibly infeasible (i) if $\beta$ is large; (ii) if the energy barriers are too high; (iii) for problems in high dimension. However, exploring data based approaches remains interesting, at least for academic reasons. 

\subsection{"Naive" Variational AutoEncoders to generate transition paths}
\label{sec:2d_vae}

We first use VAEs with a latent space of dimension 2. The encoder is composed of a convolution block denoted by CNN-A (which is a combination of some 1--dimensional convolutional layers and  batch normalization layers; see Appendix~\ref{sec:app_A} for the exact architecture).  The CNN-A is stacked with an additional linear layer that produces as output $\mu(q), \sigma(q) \in \R^2$ (which is the output of the encoder). The decoder structure is the ``transpose" structure of CNN-A (meaning that the convolutional blocks are defined with the transposed convolutional layer and the architecture parameters are the same as CNN-A but taken in the inverse order).  We used the AdamW PyTorch~\cite{loshchilov2017decoupled} optimizer with learning rate $10^{-4}$ and trained the VAE for $1400$ epochs with a batch size of $64$.  

\begin{figure}[H]
	\centering
	\begin{minipage}{0.95\textwidth}
		\centering
		\includegraphics[scale=0.6]{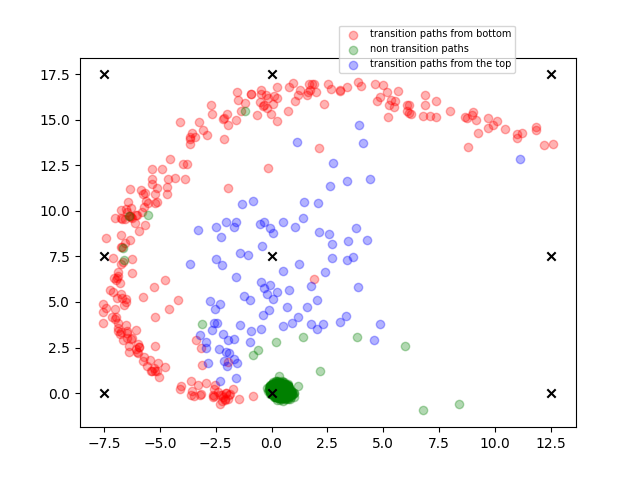}
		\caption{Mean projection of the test set onto the latent space for the "naive" VAE. The crosses represent the points used to generate new trajectories, see Figure~\ref{fig:generated_CNN_A}. }
		\label{fig:embeding_2d_CNN_A}
	\end{minipage}
\end{figure}

The results presented in Figure~\ref{fig:embeding_2d_CNN_A} provide a representation of the data in the 2-dimensional latent space corresponding to the bottleneck of the VAE. We can clearly distinguish between the three types of trajectories in this 2-dimensional space: transition paths are on the outer part of the space, while non transition paths concentrate around the origin. Once the VAE is trained, one can sample new points in the $2$--dimensional latent space, in order to generate new trajectories. 

\begin{figure}[H]
	\centering
			\begin{minipage}{0.32\textwidth}
					\centering
		\includegraphics[width=\textwidth]{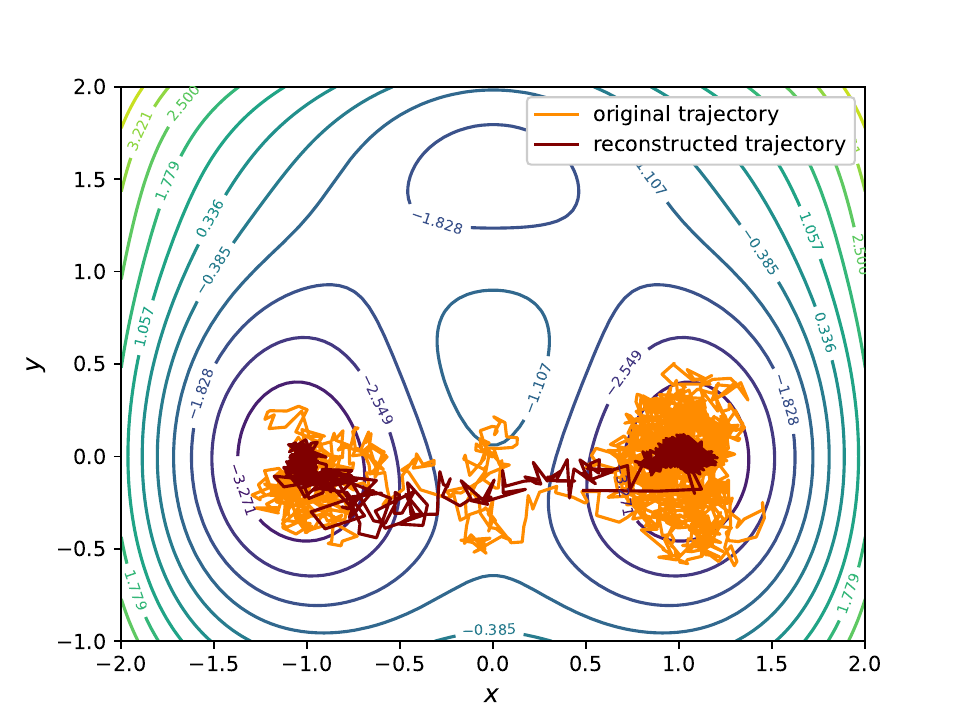}
	\end{minipage}
	\begin{minipage}{0.32\textwidth}
		\centering
		\includegraphics[width=\textwidth]{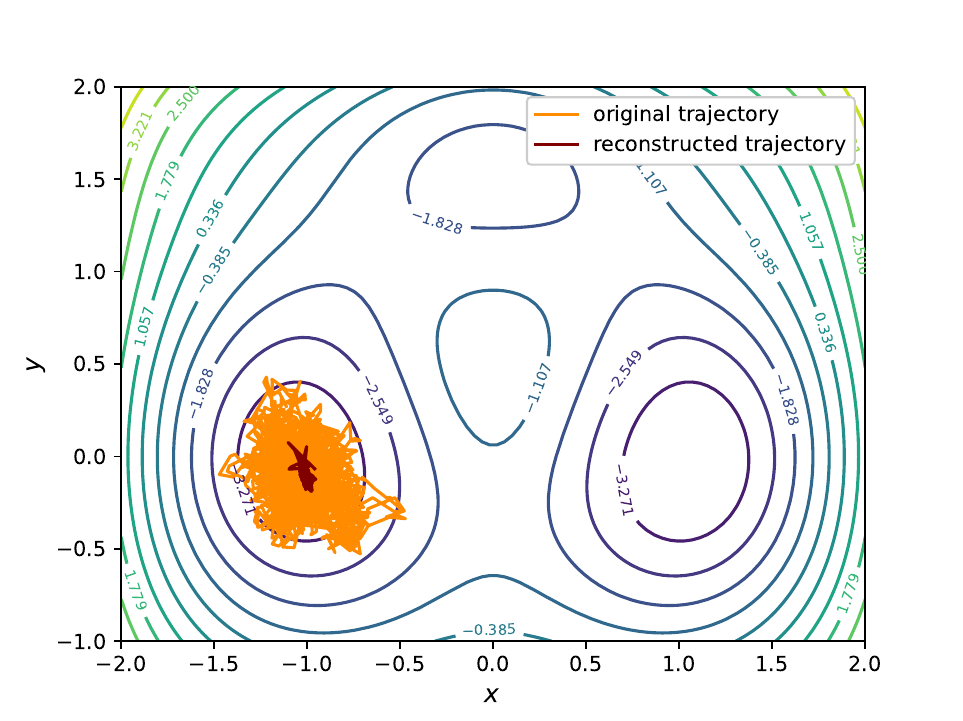}
			\end{minipage}
	\begin{minipage}{0.32\textwidth}
		\centering
	\includegraphics[width=\textwidth]{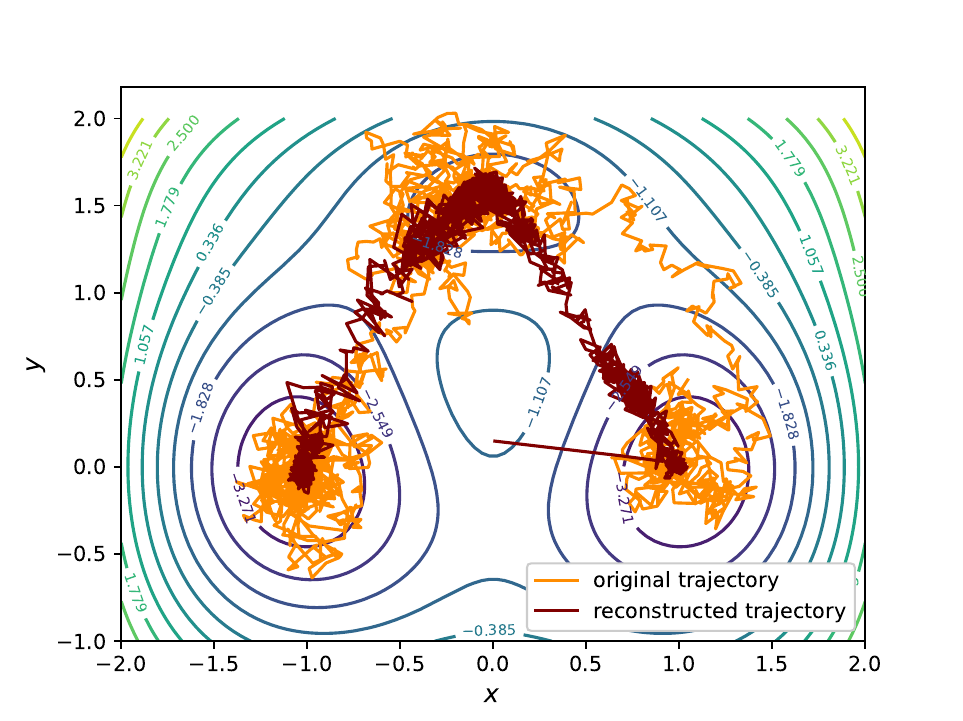}
\end{minipage}
		\caption{Comparison between original and reconstructed trajectories using the trained "naive" VAE. The orange lines represent the original trajectories (from the test set) and the brown lines the reconstructed ones. }
		\label{fig:reconstructed_CNN_A}
\end{figure}

We first plot in Figure~\ref{fig:reconstructed_CNN_A} some reconstructed trajectories and the original ones (from the test set) using the trained VAE. The trajectories reconstructed by the VAE have the correct shape but the magnitude of oscillations of the configurations are small compared to the original ones. The sampling of the well $A$ is also better in the original trajectories. 

\begin{figure}[H]
	\centering
			\begin{minipage}{0.32\textwidth}
		\centering
		\includegraphics[width=\textwidth]{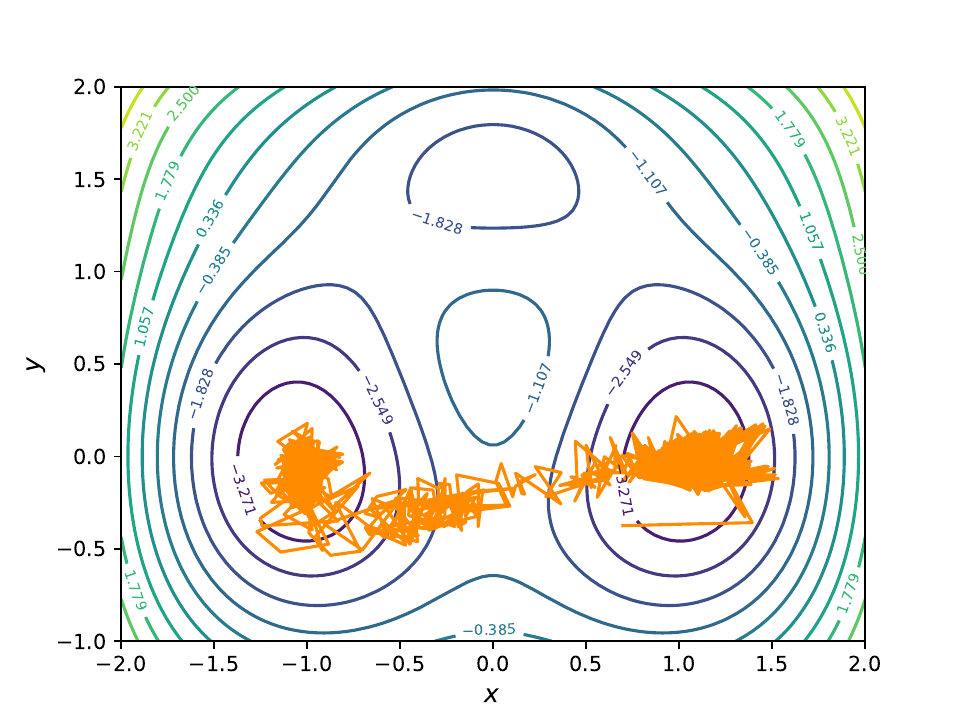}
		\subcaption{$z = (-7.5, 17.5)$}
	\end{minipage}
	\begin{minipage}{0.32\textwidth}
		\centering
		\includegraphics[width=\textwidth]{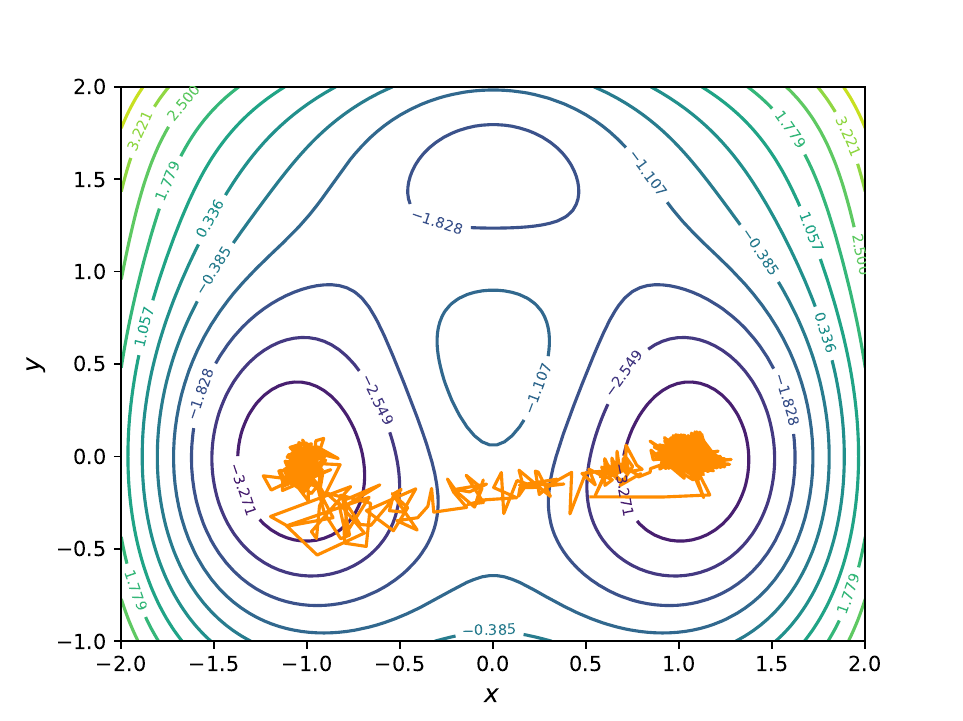}
		\subcaption{$z = (0, 17.5)$}
	\end{minipage}
	\begin{minipage}{0.32\textwidth}
		\centering
		\includegraphics[width=\textwidth]{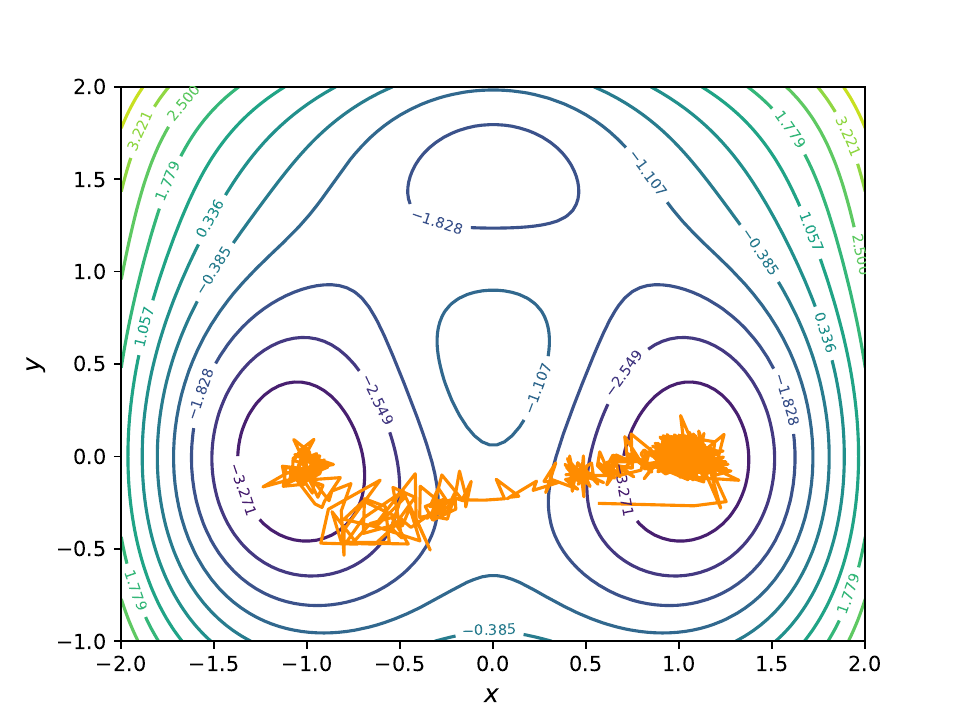}
		\subcaption{$z = (12.5, 17.5)$}
	\end{minipage}
		\begin{minipage}{0.32\textwidth}
		\centering
		\includegraphics[width=\textwidth]{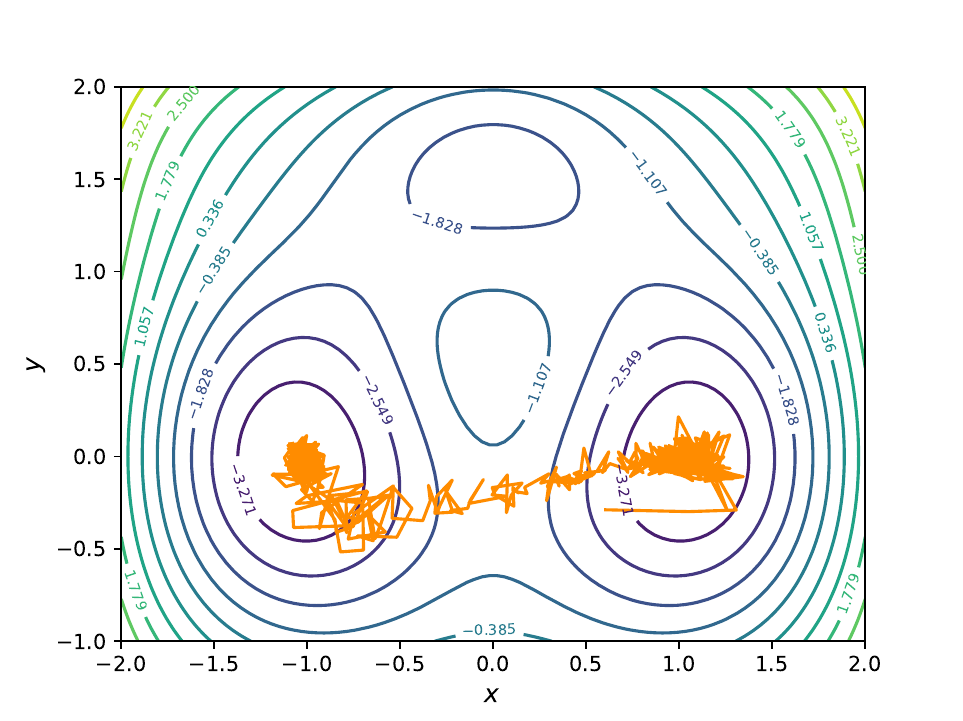}
		\subcaption{$z = (-7.5, 7.5)$}
	\end{minipage}
	\begin{minipage}{0.32\textwidth}
		\centering
		\includegraphics[width=\textwidth]{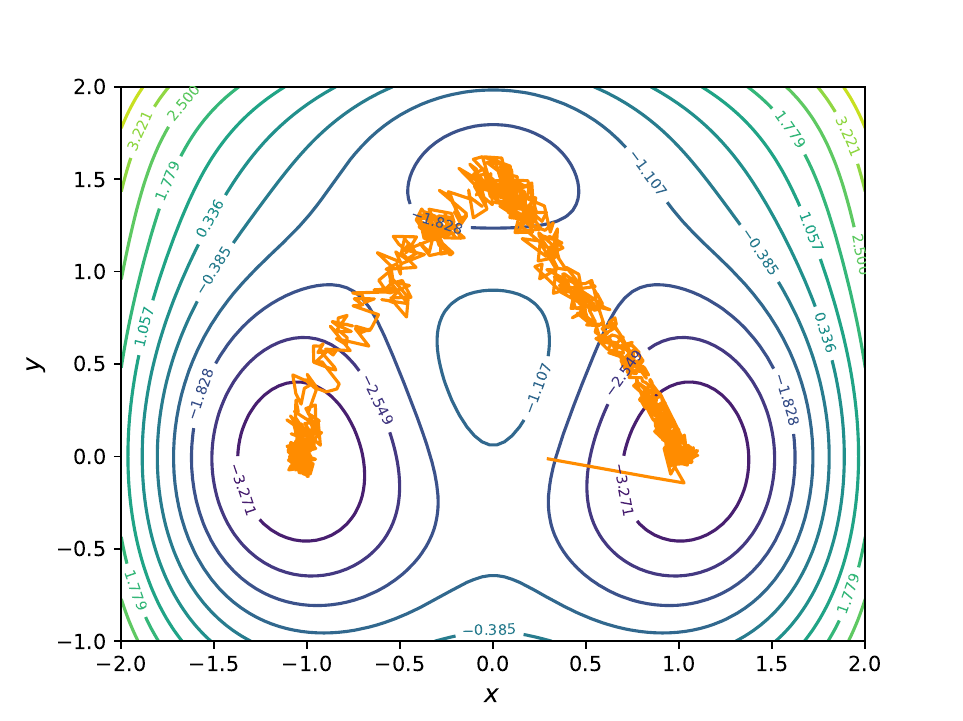}
		\subcaption{$z = (0, 7.5)$}
	\end{minipage}
	\begin{minipage}{0.32\textwidth}
		\centering
		\includegraphics[width=\textwidth]{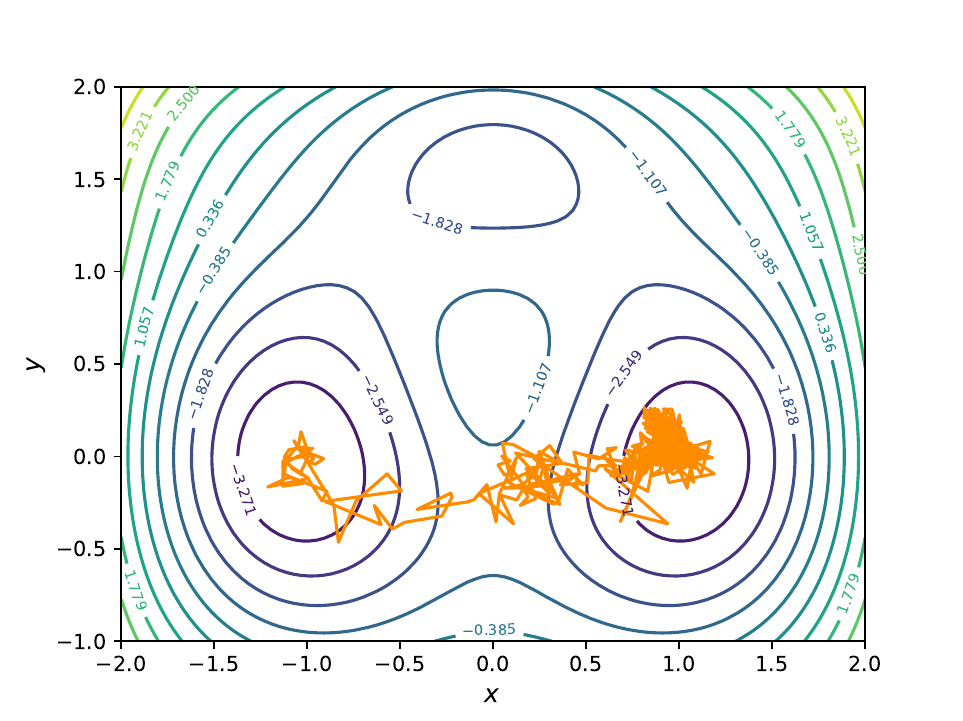}
		\subcaption{$z = (12.5, 7.5)$}
	\end{minipage}
	\begin{minipage}{0.32\textwidth}
		\centering
		\includegraphics[width=\textwidth]{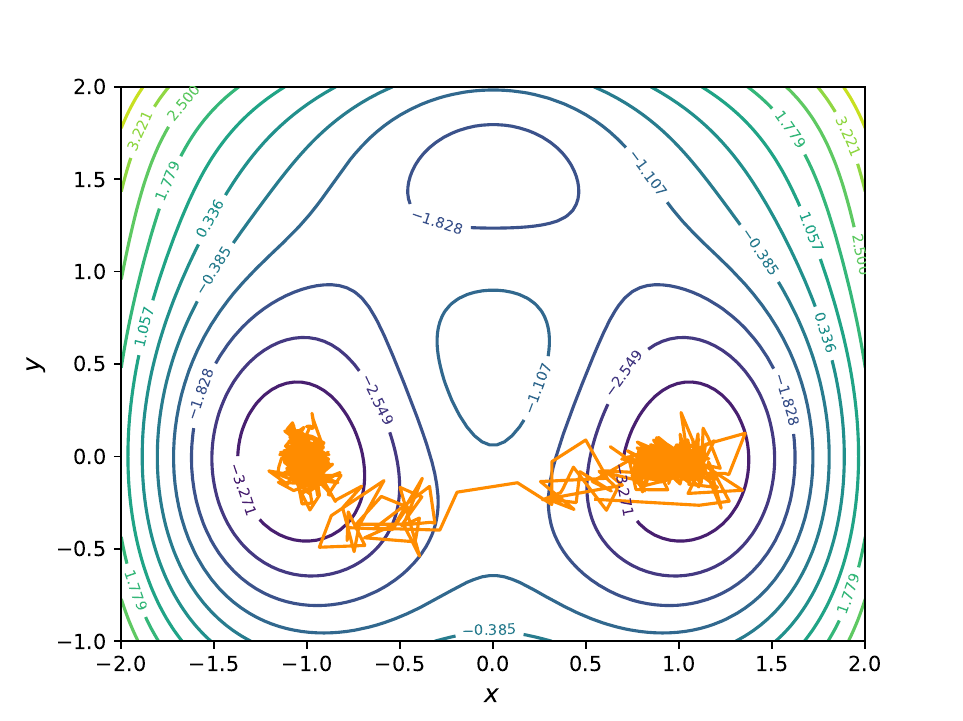}
		\subcaption{$z = (-7.5, 0)$}
	\end{minipage}
	\begin{minipage}{0.32\textwidth}
		\centering
		\includegraphics[width=\textwidth]{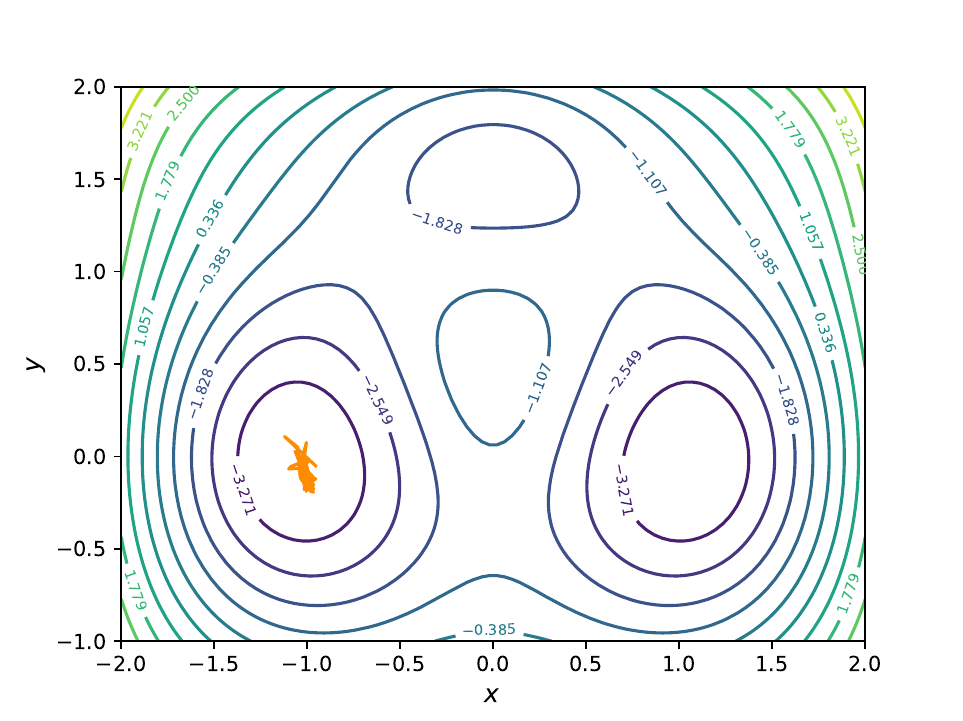}
		\subcaption{$z = (0, 0)$}
	\end{minipage}
	\begin{minipage}{0.32\textwidth}
		\centering
		\includegraphics[width=\textwidth]{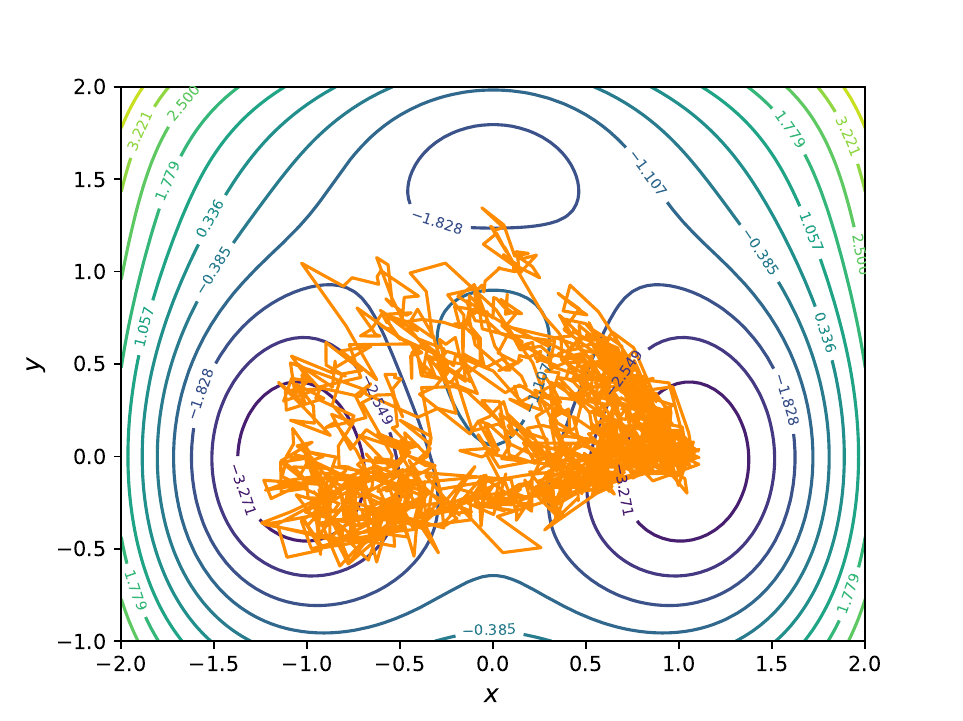}
		\subcaption{$z = (12.5, 0)$}
	\end{minipage}
	\caption{Generated trajectories using the trained "naive" VAE, for various values of the latent variables (see crosses in Figure~\ref{fig:embeding_2d_CNN_A}).}
	\label{fig:generated_CNN_A}
\end{figure}

\begin{figure}[H]
	\centering
	\begin{minipage}{0.95\textwidth}
		\centering
		\includegraphics[scale=0.45]{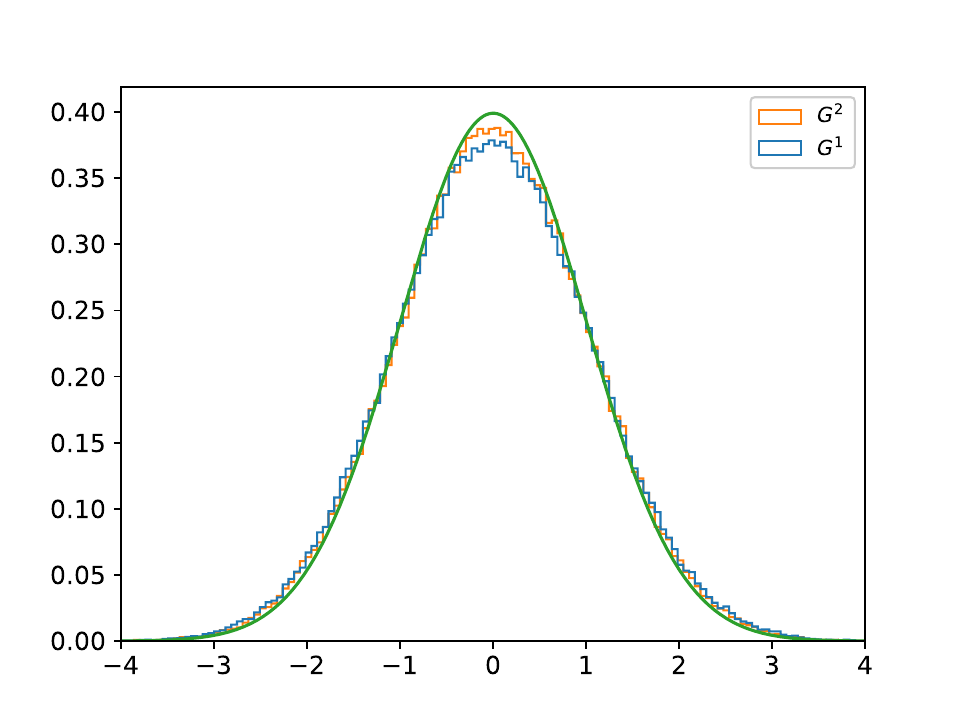}
        \includegraphics[scale=0.45]{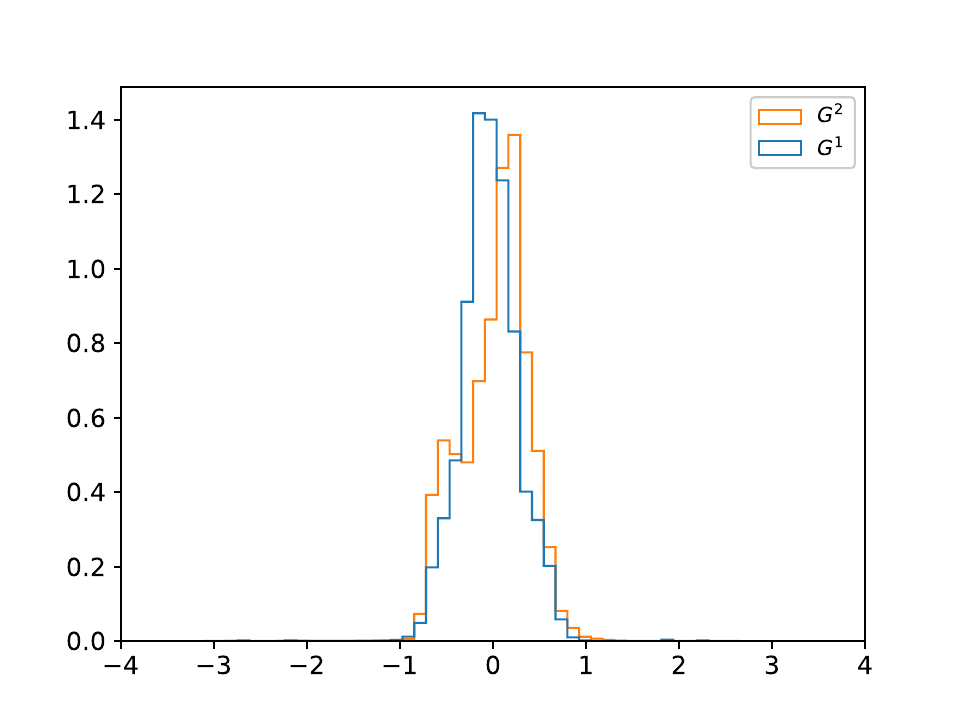}
		\caption{Distributions of the Gaussian increments over the original and generated trajectories. }
		\label{fig:gaussians_relevance}
	\end{minipage}
\end{figure}

We plot in Figure~\ref{fig:generated_CNN_A} some generated trajectories using points from the latent space represented in Figure~\ref{fig:embeding_2d_CNN_A} by crosses, in an attempt to obtain new trajectories by an extrapolation procedure in latent space. Although their overall shape is rather correct (except for some trajectories far from the data points in latent space), most of the generated trajectories are not really convincing, as they are too straight and lack the erratic motion around some mean path arising from the Brownian term in the reference dynamics~\eqref{eq:O_Langevin}. To quantify this, we compute the Gaussians increments which are needed to observe a given trajectory, by inverting~\eqref{eq:Langevin-discrete}. More precisely, for a given trajectory $q = (q_0,\dots,q_T)$, we compute 
\[
\mathcal{G}_k = \sqrt{\frac{\beta}{2\Delta t}} \left( q_{k+1} - q_k +\nabla V(q_k)\Delta t \right).
\]
The histograms of the components of the two dimensional vectors~$(\mathcal{G}_0,\dots,\mathcal{G}_{T-1})$ over 100 randomly chosen trajectories are plotted in Figure~\ref{fig:gaussians_relevance}. The distribution of the values of~$\mathcal{G}$ for the generated trajectories are much narrower than the corresponding distribution for the trajectories from the training dataset, which corresponds to a standard Gaussian distribution.
The analysis could be refined here by computing the path action, as considered in~\cite{GRG22} to assess the relevance of generated trajectories. The path action is proportional to the sum~$\mathcal{G}_0^2 + \dots + \mathcal{G}_{T-1}^2$, which, unsurprisingly here in view of Figure~\ref{fig:gaussians_relevance}, turns out to have untypical values (too small).

With the VAE architecture we consider, a complete trajectory is encoded in the very low dimensional space~$\R^2$, which is too small to account for the fluctuations of the whole trajectory, and can at best reproduce the average shape of trajectories relating two metastable states. One way to deal with this issue and reintroduce some Brownian-like fluctuation in the proposed trajectories would be to post-process the generated trajectories using transition path sampling techniques~\cite{doi:10.1146/annurev.physchem.53.082301.113146} to locally relax the generated paths, for instance by relying on the so-called Brownian tube proposal suggested in~\cite{ Stoltz07}. We take an alternative route in Section~\ref{sec:64d_vae} and try instead to incorporate the temporal aspect in the latent space by increasing the dimension of the latent space. 

\subsection{VAEs with larger latent space}
\label{sec:64d_vae}

To take  into account the temporal aspect of the data, we use a VAE composed of the same convolutional blocks denoted by CNN-A in Appendix~\ref{sec:app_A}, and an additional linear layer as encoder to produce an output $\mu(q), \sigma(q) \in \R^{T_z \times 2}$, where $T_z = T/2^6 = 31$ (because we chose to work with a neural network composed of~$6$ convolutional layers for which the length of the input is divided by~2). This means that each trajectory is encoded by a latent variable of dimension $T_z\times 2$. The inputs of the decoder are $z = (z_1, ..., z_{T_z})$, with $z_i = \mu(q)_i +  \sigma(q)_i \varepsilon _i$, where~$(\varepsilon_i)_{1 \leq i \leq T_z}$ are independent and identically distributed standard 2-dimensional Gaussian random vectors. Again, the decoder is simply the  "transpose" of the encoder. To train the model, we use again the AdamW PyTorch optimizer with learning rate $10^{-4}$ for $1400$ epochs with batch size~$64$. 

\begin{figure}[H]
\centering
\begin{minipage}{0.9\textwidth}
    \centering
    \includegraphics[scale=0.6]{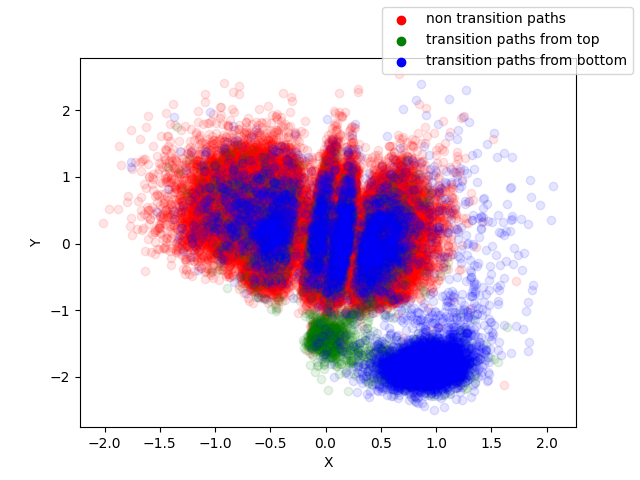}
    \caption{Mean projection of the test set onto the latent space (using the VAE with a latent space of dimension $31 \times 2$). Each trajectory is represented by $31$ points.}
    \label{fig:cemracs_cnn_embedding.png}
\end{minipage}
\end{figure}

\begin{figure}[H]
	\centering
	\begin{minipage}{0.32\textwidth}
		\centering
		\includegraphics[width=\textwidth]{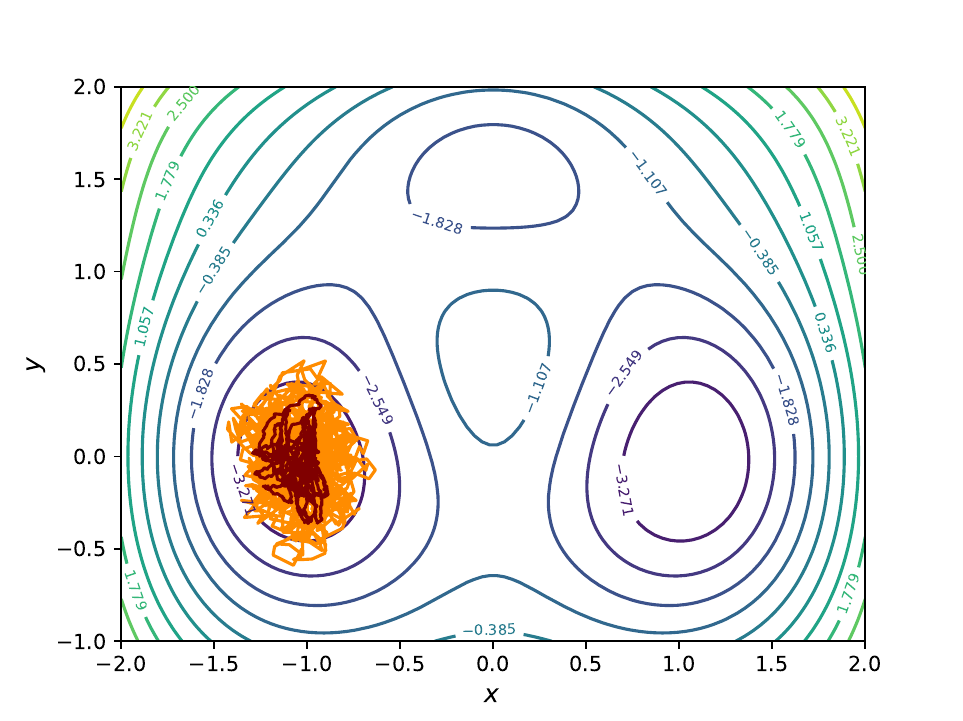}
	\end{minipage}
	\begin{minipage}{0.32\textwidth}
		\centering
		\includegraphics[width=\textwidth]{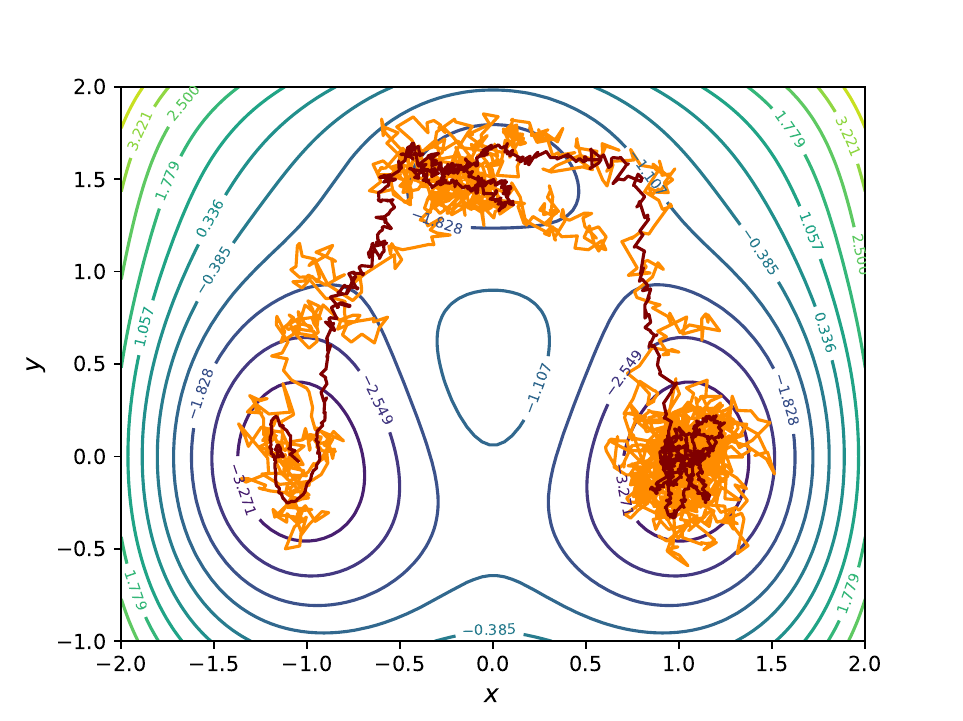}
	\end{minipage}
	\begin{minipage}{0.32\textwidth}
		\centering
		\includegraphics[width=\textwidth]{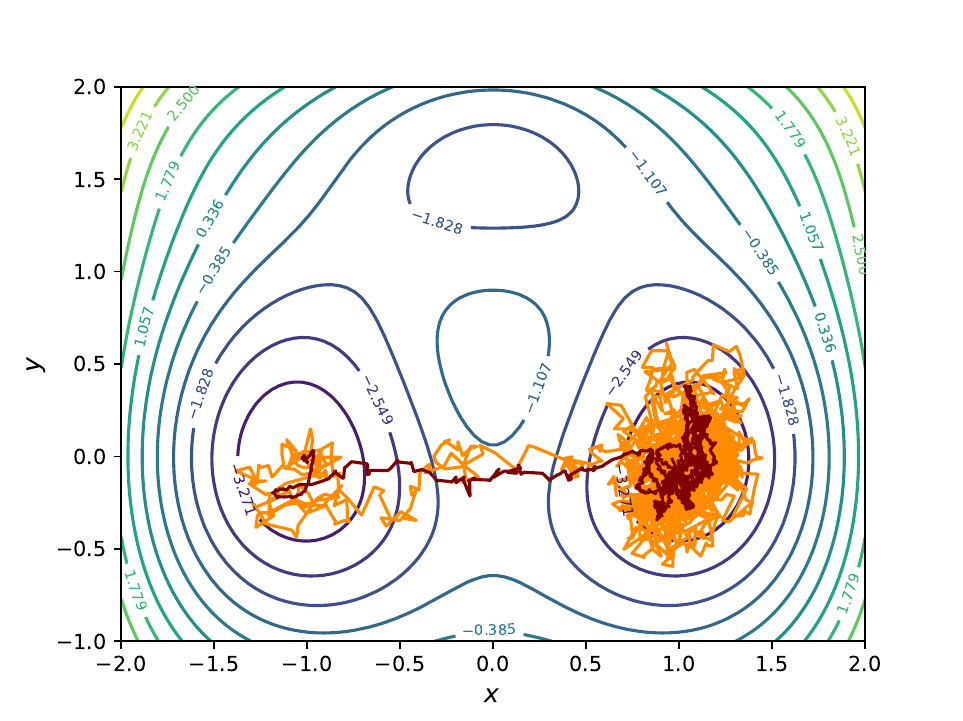}
	\end{minipage}
	\caption{Comparison between original and reconstructed trajectories using the trained VAE with a latent space of dimension~$31$. The orange lines represent the original trajectories (from the test set) and the brown lines the reconstructed ones.}
	\label{fig:cemracs_generated_trajectoires_cnn}
\end{figure}

We first plot in Figure~\ref{fig:cemracs_cnn_embedding.png} the mean projections onto the latent space for the test set, corresponding to the scatter plot of the elements of the vectors $\mu(q)$ for each trajectory $q$ in the test set. We can see the zone indicating the windows that are in~$A$ or in~$B$, as well as the reactive parts of the trajectory which corresponds to the part of the transition path between the last time it leaves~$A$ and the first time it enters~$B$. We plot in Figure~\ref{fig:cemracs_generated_trajectoires_cnn}  the trajectory reconstructions when passed through our VAE architecture for trajectories in the test set. The results are better than the ones obtained in Section~\ref{sec:2d_vae}. The fluctuations in the generated trajectories are more representative of the ones of the original trajectories. The problem with such a model is that we cannot generate new trajectories since we would need to this end a meaningful sequence $(z_1, ... z_{T_z})$ in the latent space. The difficulty is that the distribution of such sequences on the latent space obtained by an application of standard VAEs does not incorporate temporal information on the initial trajectory, for instance taking into account correlations between subsequent values of the variables~$z_i$.

In order to do so, we tried a very recent development for learning tasks on data with a recursive structure (be it one dimensional time series as audio or two dimensional data like images): the vector quantized variational autoencoder network (VQ-VAE)~\cite{oord2018neural}. The main idea of this method is to first learn a quantization of our feature space $z$ so that each input trajectory $q_1, ..., q_T$ is represented by $z_1, ..., z_{T_z}$ where $z_i$ is an element of a finite set of  vectors $Z = \{\tilde{z}^1,..., \tilde{z}^K\} \in \R^\ell$, which is also learned by the autoencoder. Once a good quantized autoencoder is available, it can be used to learn an autoregressive model, defined on the finite set of possible elements of~$Z$. More precisely, denoting by~$I_{i}$ the index of the projection for the part~$i$ of the sequence represented in~$Z$, namely~$z_i = \tilde{z}^{I_i}$, the autoregressive model learns a transformation predicting~$I_k$ from~$I_1, ..., I_{k-1}$. The trajectories generated with VQ-VAE were however not convincing, which is why we do not report their results here. We also considered introducing a variational recurrent neural network~\cite{chung2016recurrent} between the encoder and the decoder to learn the distribution on the latent space, but this also led to poor performances when generating trajectories. This motivates moving to a data free approach, as we do in Section~\ref{sec:RL}.

\section{Generating transition paths with reinforcement learning}
\label{sec:RL}

We turn in this section ta data-free learning approaches based on reinforcement learning. The idea is to learn a drift term to bias the dynamics and force the system to go from one metastable to the other. We first present the main setting of reinforcement learning alongside with the Q-learning framework in Section~\ref{sec:main_set}. We next present in Section~\ref{sec:app_RL} how we apply it to our problem, and make precise in particular the reward function used to train the models. Finally, we present in Section~\ref{sec:num_RL} the numerical results obtained with this approach. 

\subsection{Overview of reinforcement learning}\label{sec:main_set}

The conceptual framework behind reinforcement learning is the theory of Markov decision processes. A Markov decision process is defined by a 4-tuple ($\mathcal{Q}$, $\mathcal{A}$, $\mathcal{T}$, $r$) with 
\begin{itemize}
\item  a set of states $\mathcal{Q}$;
\item a set of actions $\mathcal{A}$;
\item a family of transition kernels $\mathcal{T}(q_{k+1} | q_k, a_k)$ where $q_{k}, q_{k+1} \in \mathcal{Q}$ and $ a_k \in \mathcal{A}$;
\item   a reward function $r_{a_k}(q_{k+1}, q_k)$ where $ q_k, q_{k+1} \in \mathcal{Q}$ and $  a_k \in \mathcal{A}$.
\end{itemize}
This framework allows to simulate  a "game" consisting of a set of states $\mathcal{Q}$, where at each step an action~$a_k$ is selected depending on the current state~$q_k$ of the system. This action $a_k$ leads to a transition to the state $q_{k+1}$ with probability $\mathcal{T}(q_{k+1} | q_k, a_k)$, with an associated reward $r_{a_k}(q_{k+1}, q_k)$. In this work, actions are selected according to a deterministic policy function $P: \mathcal{Q} \rightarrow \mathcal{A}$.  For a given policy~$P$, an important function is the value function~$V_P$, which gives the (discounted) cumulated expected reward obtained when starting from the position $q_k$ at time~$k$ when the agent follows the policy $P$: \begin{equation}\label{eq:goal_rl}
    V_P(q_k) =\mathbb{E}_{\mathcal{T}_{P}} \left[\sum_{i=k}^\infty \gamma^{i-k} r_{P(q_i)}(q_{i+1}, q_i)\right],
\end{equation}
In the latter expression, expectations are taken with respect to realizations of the Markov chain with transition kernel~$\mathcal{T}_P(q_{k+1}|q_{k}) := \mathcal{T}(q_{k+1} | q_{k}, P(q_{k}))$, and $\gamma \in (0, 1)$ is a discount factor, which balances the importance between having a gain at time step~$k$ and having a gain in the future.
We also recall the action value function (also called the Q-function), which returns the average expected reward of taking an action $a_{k}$ from state $q_{k}$ and then following the policy~$P$. The action value function reads
\begin{equation}\label{eq:q_function_def}
     Q_P (q_{k}, a_{k}) = \mathbb{E}_{\mathcal{T}(\cdot|q_{k},a_{k}) \otimes \mathcal{T}_P} \left[r_{a_{k}}(q_{k+1}, q_{k} )+ \sum_{i=k+1}^\infty \gamma^{i-t} r_{P(q_i)}(q_{i+1}, q_{i})\right],
\end{equation}
where the expectation is taken over realizations of~$q_{k+1}$ distributed according to~$\mathcal{T}(\cdot|q_{k},a_{k})$, and~$q_{i+1} \sim \mathcal{T}_P(\cdot|q_i)$ for~$i \geq k+1$. In particular, $V_P(q_{k}) = Q_P(q_{k},P(q_{k}))$. The optimal value function is defined for each state~$q_{k}$ by finding a policy which maximizes the value function (in particular, this policy depends a priori on the state~$q_{k}$):
\begin{equation}\label{eq:goal_rl}
    V^*(q_{k}) = \max_{P: \mathcal{Q} \rightarrow \mathcal{A}} V_P(q_{k}), 
\end{equation}
Similarly, the optimal action value function is defined for each state-action pair $(q_{k}, a_{k})$ as 
\begin{equation}\label{eq:goal_rl}
    Q^*(q_{k}, a_{k}) = \max_{P: \mathcal{Q} \rightarrow \mathcal{A}} Q_P(q_{k}, a_{k}).
\end{equation}

An important issue at this stage is to find a policy which is optimal in some sense. A natural way to define a partial order on the policies is the following: $P_1 \geq P_2$ if $V_{P_1}(q) \geq V_{P_2}(q)$ for any~$q \in \mathcal{Q}$. An optimal policy~$P^*$ is  therefore such that $V_{P^*}(q) \geq V_P(q)$ for any~$q \in \mathcal{Q}$. One can prove that $P^*$ given by
\begin{equation}
\label{eq:pi_star_as_argmax}
    P^*(q) = \argmax_{a \in \mathcal{A}} Q^*(q, a),
\end{equation}
is an optimal policy~\cite{puterman2014markov}. Furthermore, $V^* = V_{P^*}$ and $Q^* = Q_{P^*}$, see again~\cite{puterman2014markov} for a proof. One way to determine the optimal policy is then to solve the problem~\eqref{eq:pi_star_as_argmax}, using the expression~\eqref{eq:goal_rl} for~$Q^*(q_{k}, a_{k})$. 

\paragraph{Q-Learning.} When the action space is finite, a common approach for solving the optimization goal stated in \eqref{eq:goal_rl} is to rely on the Hamilton--Jacobi--Bellman formulation of the Q-function, directly obtained from~\eqref{eq:q_function_def}:
\begin{equation}\label{eq:q_hjb}
    Q_P (q_{k}, a_{k}) = \mathbb{E}_{\mathcal{T}(\cdot|q_{k},a_{k})} \left[r_{a_k}(q_{k+1}, q_{k}) + \gamma Q_P(q_{k+1}, P(q_{k+1})) \right],
\end{equation}
where the expectation is over all realizations of~$q_{k+1}$ distributed according to~$\mathcal{T}(\cdot|q_k,a_k)$. For a discrete action space $\mathcal{A}$, the algorithm for Q learning is based on the fact that $Q_P$ can be seen as a fixed point of the mapping appearing in~\eqref{eq:q_hjb}. We briefly describe the method following~\cite{qlearning}. An initial Q-table is considered, for instance by providing random values for the Q-function for each pair~$(q_{k}, a_{k})$. At each time step $k$, the agent takes an action $a_k$ (either randomly or by choosing the best one among the current estimates for the Q function, see~\cite{qlearning} for more details) and moves from the state $q_k$ to $q_{k+1}$ with a reward $r_{a_k}(q_{k+1}, q_k)$. In view of~\eqref{eq:q_hjb}, the Q-table is then updated as follows
\[ Q(q_k, a_k) \leftarrow (1- \alpha) Q(q_k, a_k)  + \alpha \left(r_{a_k}(q_{k+1}, q_k)   + \gamma \max_{a \in \mathcal{A}} Q(q_{k+1}, a)  \right) ,\]
where $\alpha$ is the learning rate. The convergence of Q-learning algorithms can be shown under mild hypotheses, see~\cite{qlearning, sutton2018reinforcement} for more details. Once the algorithm has converged and a final Q-function is obtained, the optimal policy at state~$q_{k}$ is defined as $\argmax_{a \in \mathcal{A}} Q(q_{k}, a)$. 

In large state spaces, either discrete or continuous, a better way to maximize~\eqref{eq:q_function_def} is to consider a family of functions $f(q_{k}, a, \theta)$ parametrized by~$\theta$ to approximate the Q-function. This is more efficient in terms of computational time if the state space is discrete but large. Typically, $\theta$ are the parameters of a NN. The policy in this case would be $P_\theta(q_{k}) = \argmax_{a \in \mathcal{A}} f(q_{k}, a, \theta)$. In practice, one iterates between updating a data set of realizations (here, snapshots of the trajectory instead of full trajectories, in fact; see Section~\ref{sec:app_RL} below), and the approximation of the Q function via~$f(q_{k},a,\theta)$, in an online manner. More precisely, at each iteration, the algorithm consists in adding to the data set a new observed transition from~$q_{N-1}$ to~$q_N$ when performing the action~$a_N$:
\[
D_N = D_{N-1} \cup \{(q_N, a_N, q_{N+1}) \}, \qquad D_0 = \emptyset,
\]
and updating the parameters of the NN based on the new data set~$D_N$ as
\begin{equation}\label{eq:q_learning_opt}
\theta^* = \argmin_{\theta'} \sum_{i=1}^N\left[ f(q_i, a_i, \theta') - \left(r_{a_i}(q_{i+1}, q_i) + \gamma \max_{a \in \mathcal{A}} f(q_{i+1}, a, \theta')\right) \right]^2.
\end{equation} 
The precise procedure, as well as various strategies to make the numerical method more stable, can be read in~\cite{mnih2013atari}.


\paragraph{Infinite action space.} When the action space is continuous, taking the maximum over the action space is impossible. One way to solve the Q-learning task, is to introduce an additional neural network, the policy network~$\mathcal{P}(q_{k}, \omega)$, which aims at learning the optimal policy~$P^*(q_{k})$. As the policy network is the one proposing the action to be taken, it is called an ``actor''. The policy network is used in conjunction with another neural network~$\mathcal{Q}(q_{k}, a_{k}, \theta)$ which evaluates the performance of the action which was taken by approximating the Q-function. The function~$\mathcal{Q}$ is the counterpart for continuous action spaces of the function~$f$ introduced above for discrete action spaces. As the network~$\mathcal{Q}$ evaluates the actor, it is called a ``critic'', and the whole learning procedure forms an ensemble called the ``actor-critic setting''.  From a mathematical viewpoint, one aims at solving the following optimization problem:
\begin{align}
    \omega^* &= \argmax_{\omega} \sum_{i=1}^N \mathcal{Q}(q_i, \mathcal{P}(q_i,  \omega), \theta^*) \label{eq:actor_optimization},\\
    \theta^* &= \argmin_{\theta}  \sum_{i=1}^N \left[\mathcal{Q}(q_i, \mathcal{P}(q_i,  \omega^*), \theta) - \left(r_{a_i}(q_{i+1}, q_i) + \gamma \mathop{\mathrm{max}}_{a\in \mathcal{A}} \mathcal{Q}(q_{i+1}, a, \theta) \right)\right]^2.\label{eq:critic_optim}
\end{align}
Note that the policy network implicitly appears in~\eqref{eq:critic_optim} through the generation of the dataset. 

The maximizations in~\eqref{eq:actor_optimization}-\eqref{eq:critic_optim} are more challenging than the corresponding maximization problem~\eqref{eq:q_learning_opt} for finite action spaces. Naive strategies are known to be ill behaved, leading notably to overconfident predictions on the $\mathcal{Q}$ networks (\emph{i.e.} the resulting approximation of the Q-function gives values higher than the actual ones). Several ways of dealing with this issue have been proposed over the years. We follow the so called Twin Delayed Deep Deterministic policy gradient (TD3) algorithm defined in~\cite{TD3}.

This algorithm introduces several tools to stabilize the learning of~$\mathcal{Q}$ and~$\mathcal{P}$. One of the main ideas is to replicate the $\mathcal{Q}$ network into two $\mathcal{Q}$ networks with parameters $\theta_1, \theta_2$. Taking the minimal value of $\mathcal{Q}_{\theta_1}(q, a)$ and $\mathcal{Q}_{\theta_2}(q, a)$ allows to mitigate the issue of overconfident $\mathcal{Q}$ networks mentioned above. Another important idea is to use delayed networks that are copies of $\mathcal{Q}$ and $\mathcal{P}$ but in which the weights are updated with some memory function (as an exponentially weighted linear combination of current and past weights). We refer to the original paper~\cite{TD3} for a more in-depth discussion of these points.

\subsection{Application to sampling transition paths}\label{sec:app_RL}

We now describe how to adapt the reinforcement learning framework previously described to the problem of sampling transition paths described in Section~\ref{sec:sampling_pres}. Although we consider finite trajectories of given length, we still rely on the stationary framework introduced in the previous section. This is a valid approximation when trajectories are sufficiently long. 

We successively define all the elements of the Markov decision process introduced in Section~\ref{sec:main_set}.

\paragraph{State space and action space.} The state space is simply defined as the space $\mathcal{D}$ of the diffusion process, $\mathcal{D} \subset \R^d$. The action space is the same space as the image of the gradient of the potential, \emph{i.e.}~$\mathbb{R}^d$. In this context, we consider the policy to be a vector field, introduced as a bias into the governing equation~\eqref{eq:O_Langevin} to alter the trajectory. This is similar to the choice made in~\cite{schutte2012optimal, hartmann2012efficient} where controlled stochastic differentials are considered. More explicitly, the controlled version of~\eqref{eq:O_Langevin} we consider reads
\begin{equation}
    \label{eq:O_Langevin_altered}
    dq_t = (-\nabla V(q_t) + P(q_t))\,dt + \sqrt{\frac{2}{\beta}}\,dW_t,
\end{equation}
discretized in the actor-critic framework as
\begin{equation}
    \label{eq:Langevin_altered-discrete}
    q_{k+1} = q_k +(-\nabla V(q_k)+\mathcal{P}(q_k,\omega))\Delta t + \sqrt{\frac{2\Delta t}{\beta}}G_k,
\end{equation}
where~$\omega$ denotes the state of the policy network. We chose here to work with a generic policy, although we could have looked for it in gradient form according to results of stochastic optimal control (as reviewed in~\cite[Section~6.2]{lelievre_stoltz_2016}). We also denote by~$\mathcal{T}_\mathcal{P}$ the corresponding transition kernel in the sequel (not explicitly writing out the state~$\omega$ of the neural network).

\paragraph{Probability kernel.} Given a policy~$\mathcal{P}(\cdot,\omega)$, the probability kernel to go from a configuration $q_{k}$ to a new one $q_{k+1}$ can be deduced from~\eqref{eq:Langevin_altered-discrete} to have the density
\begin{equation}
    \mathcal{T}_\mathcal{P}(q_{k+1} | q_{k}) = \left(\frac{\beta}{4 \pi \dt}\right)^{d/2} \exp\left( -\beta \frac{\|q_{k+1} - q_{k} - \dt(-\nabla V(q_{k})+\mathcal{P}(q_k,\omega))\|^2}{4\dt} \right).
\end{equation}

\paragraph{Reward function.}
We want to maximize the likelihood of trajectories leaving the well $A$. The reward function we consider to this end reads (omitting the dependence on the action in the notation)
\begin{equation}
r(q_{k+1}, q_{k}) = \log\left(\frac{\mathcal{T}_0(q_{k+1} | q_{k})}{\mathcal{T}_\mathcal{P}(q_{k+1} | q_{k})} \right) + \alpha h(q_{k}, q_0),
\end{equation}
where~$\alpha \geq 0$ and~$h(\cdot, q_0)$ is a function which measures the distance of the current configuration to the center~$q_0$ of the well~$A$; for instance, $h(q_k,q_0) = \| q_k-q_0\|^2$. The first term of the reward function compensates for the bias introduced in the dynamics by computing the relative likelihood with respect to an unbiased evolution. The second term of the reward function forces the particle to leave the well~$A$. 
A simple computation gives
\begin{equation}
\begin{aligned}
\log\left(\frac{\mathcal{T}_0(q_{k+1} | q_{k})}{\mathcal{T}_\mathcal{P}(q_{k+1} | q_{k})} \right) & = \frac{\beta}{4 \dt} \left[\|q_{k+1}  - q_{k}  - \dt (-\nabla V (q_{k} )+ \mathcal{P}(q_{k}, \omega  ))\|^2 - \|q_{k+1}  - q_{k} + \dt \nabla V (q_{k} )\|^2\right]\\
& = \frac{\beta}{4 \dt} \left[-2\dt(q_{k+1}  - q_{k} ) \cdot  \mathcal{P}(q_{k}, \omega ) + \dt^2 (\|\mathcal{P}(q_{k}, \omega  )\|^2 - 2 \nabla V(q_{k}) \cdot \mathcal{P}(q_{k}, \omega  ))  \right].
\end{aligned}
\end{equation}
Using~\eqref{eq:Langevin_altered-discrete}, 
\begin{equation}
\begin{aligned}
\log\left(\frac{\mathcal{T}_0(q_{k+1} | q_{k})}{\mathcal{T}_\mathcal{P}(q_{k+1} | q_{k})} \right)  
& =  \frac{\beta}{2} \left[ -\sqrt{2 \dt \beta^{-1}}G_k \cdot \mathcal{P}(q_{k}, \omega )  - \dt^2 \|\mathcal{P}(q_{k}, \omega )\|^2  \right].
\end{aligned}
\end{equation}
Note that, when summing up over~$k$ the contributions from the transitions between~$q_k$ and~$q_{k+1}$, one ends up with a discretization of the logarithm of the Girsanov weight allowing to compare the path probabilites between two dynamics differing in their drifts. 

\begin{remark}
Other choices for the function $h$ can be considered. In particular, we tried $h(q_k, q_0) = \log(\|q_k- q_0\|^2)$ but the results were less convincing than with the choice $h(q_k, q_0) = \|q_k- q_0\|^2$. Other works consider reward functions favoring that the final state is in~$B$, rather favoring exits out of~$A$ as we do, but this requires knowing in advance the target metastable region~\cite{DRGL21}.
\end{remark}

\subsection{Numerical results}\label{sec:num_RL}
We trained neural networks using the TD3 Algorithm~\cite{TD3} to generate parameters~$(\theta_{\rm f}, \omega_{\rm f})$ which are approximations of the optimal parameters~$(\theta^*,\omega^*)$ in~\eqref{eq:actor_optimization}-\eqref{eq:critic_optim}. We use the open source code provided by the authors of~\cite{TD3}, available at {\tt https://github.com/sfujim/TD3}. We fix $\dt = 5\times 10^{-3}$ and $\alpha = 0.071$. Precise information on hyper parameters and network architectures can be found in Appendices~\ref{sec:app_B} and~\ref{sec:app_C}.

\begin{figure}[H]
	\centering
	\begin{minipage}{0.45\textwidth}
		\centering
		\includegraphics[width=\textwidth]{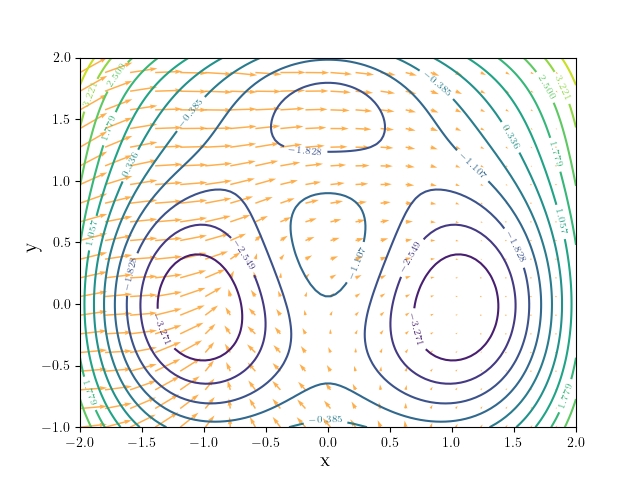}
		\end{minipage}
			\begin{minipage}{0.45\textwidth}
		\includegraphics[width=\textwidth]{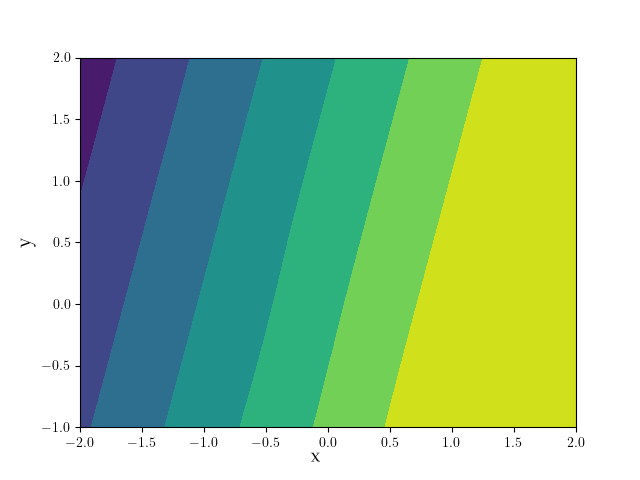}
	\end{minipage}
		\caption{Left: Visualization of the policy~$\mathcal{P}(q,\omega_\mathrm{f})$. The action field "pushes" the simulation from A through the upper saddle point and diminishes in magnitude in the neighborhood of the second metastable region. Right: Visualization of the critic value at a given position for the optimal action, namely $\mathcal{Q}(q,\mathcal{P}(q,\omega_\mathrm{f}),\theta_\mathrm{f})$.}
		\label{fig:action_field_rl}
\end{figure}

The final actor network obtained is depicted in Figure~\ref{fig:action_field_rl} where we plot the final policy~$\mathcal{P}(q,\omega_\mathrm{f})$ alongside with the final value function. We can see that the drift corresponding to the policy network biases the trajectories towards the saddle point of the potential in the vicinity of~$(0, 1)$. It does not bias the trajectories towards the saddle point around~$(0, -0.25)$, and even discourages them from performing a transition from~$A$ to $B$. This therefore biases transitions towards transitions through the upper channel. Note also that the closer the current state is to the well~$B$, the smaller the drift is, and therefore the closer the evolution is to the true dynamics. We were not able to produce transitions through the bottom saddle point.

\begin{figure}[H]
	\centering
		\begin{minipage}{0.45\textwidth}
					\centering
		\includegraphics[width=\textwidth]{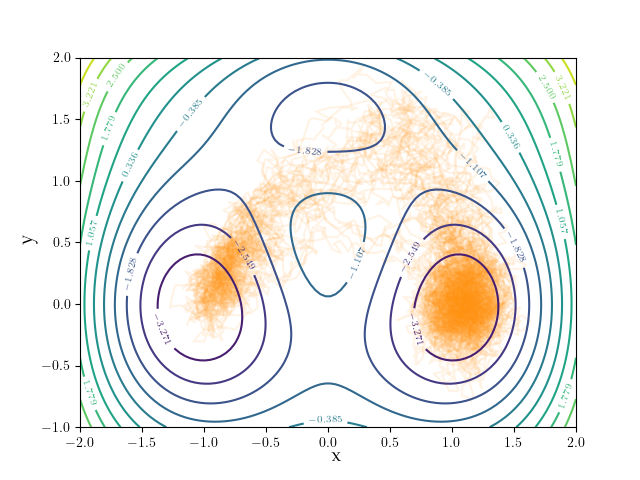}
	\end{minipage}
			\begin{minipage}{0.45\textwidth}
					\centering
		\includegraphics[width=\textwidth]{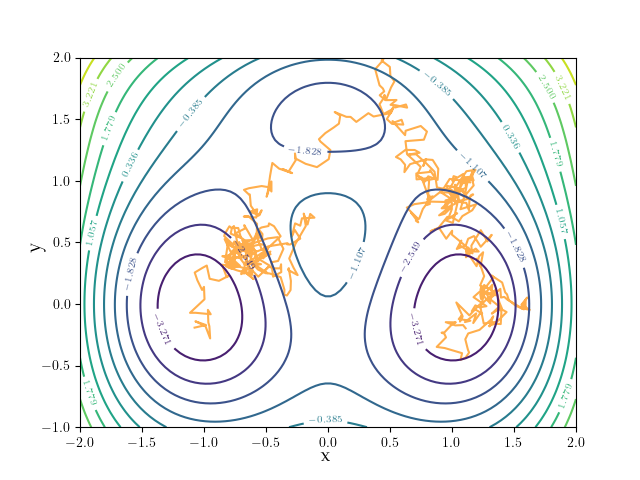}
	\end{minipage}
	\begin{minipage}{0.45\textwidth}
		\centering
		\includegraphics[width=\textwidth]{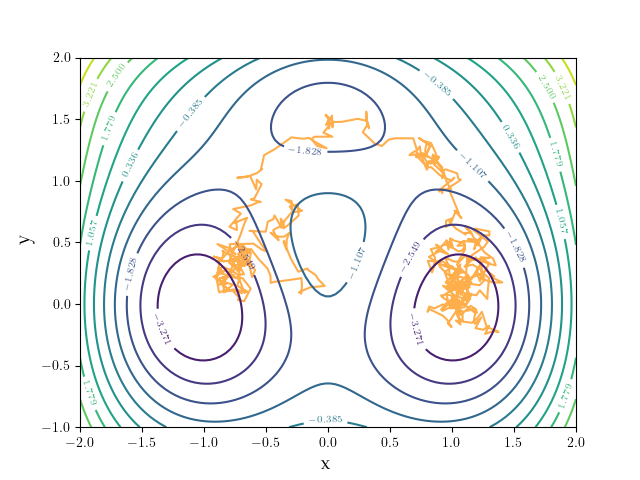}
			\end{minipage}
	\begin{minipage}{0.45\textwidth}
		\centering
	\includegraphics[width=\textwidth]{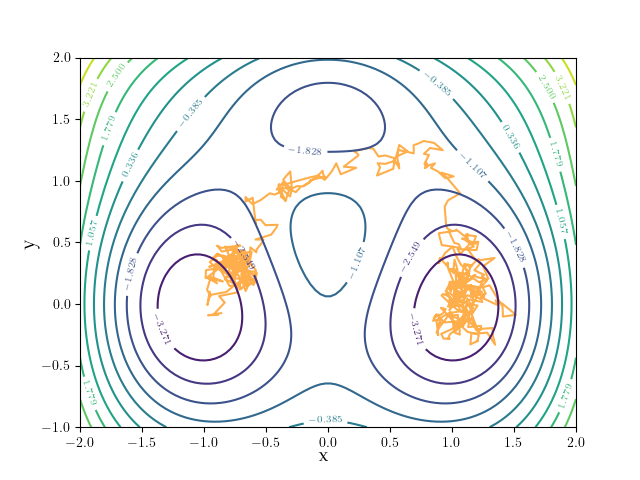}
\end{minipage}
		\caption{Sample trajectories generated using \eqref{eq:Langevin_altered-discrete} (one per picture, except the top left in which $25$ trajectories are represented).}
		\label{fig:sampled_trajectories_rl}
\end{figure}

In order to generate new transitions, we initialize the system at~$(-1, 0)$ and then run the discrete dynamics~\eqref{eq:Langevin_altered-discrete} using the final policy $\mathcal{P}(\cdot,\omega_{\rm f})$ for~$600$ time steps with~$\dt = 5\times 10^{-3}$. We generated $1000$ trajectories, and observed transitions from one metastable state to the other for more than $99\%$ of the trajectories. Some of these trajectories are plotted in Figure~\ref{fig:sampled_trajectories_rl}. They are visually more realistic than those produced in Figures~\ref{fig:reconstructed_CNN_A} and~\ref{fig:cemracs_generated_trajectoires_cnn}. Another advantage of the method is that, once the network is trained, there is no need to sample a latent variable to generate a new trajectory, as for example in Figure~\ref{fig:cemracs_generated_trajectoires_cnn}.

\section{Discussion and perspectives}

The results presented in this work suggest that reinforcement learning provides a way to sample transition paths by finding some biasing force field guiding the system in its excursion out of~$A$ and into~$B$. Approaches based on generative methods such as variational autoencoders are intrinsically more limited, especially given that they need a database of transition paths to start with. Even when such a database is available, the generation of new paths may be cumbersome as this requires a large latent space with some structure to reproduce the temporal organization of the components of the latent variables. 

While writing up this work, we became aware of recent works in the computational statistical physics community making use of reinforcement learning and neural networks to construct effective biases and favor otherwise unlikely transitions, which is particularly useful when computing large deviations rate functionals~\cite{WT20,RMG21,DRGL21,YTR22,GRG22,CTW22}. We believe that further efforts are required to better understand various choices in the reinforcement learning procedure, in particular the reward function. From a theoretical perspective, this calls for a better understanding of the links between reinforcement learning and optimal importance sampling in path space (see for instance~\cite[Section~6.2]{lelievre_stoltz_2016} and references therein).

\paragraph{Acknowledgements.} We thank Yannis Pantazis (FORTH) for stimulating discussions and preliminary attempts on the generation of transition paths in molecular dynamics; as well as Wei Zhang (FU Berlin) and the referees for their relevants comments on the initial version of this work. I.S. gratefully acknowledges financial support from Université Mohammed VI Polytechnique. The work of T.L. and G.S. is funded in part by the European Research Council (ERC) under the European Union's Horizon 2020 research and innovation programme (grant agreement No 810367).

\appendix

\section{Architecture of CNN-A used in Section~\ref{sec:vae}}\label{sec:app_A}

The following sequence of block parameters was used for the encoder network CNN-A, for an input of size~$T \times d$ with~$d=2$:
\begin{itemize}
\item layer 1: $m_{\mathrm{in} }^1=2$, $m_{\mathrm{out} }^1 = 30$, kernel size $k=4$, stride $s=2$, $T_\mathrm{out}^1 = T/2$
\item batch normalization layer
\item layer 2: $m_{\mathrm{in} }^2=30$, $m_{\mathrm{out} }^2 = 20$, kernel size $k=4$, stride $s=2$, $T_\mathrm{out}^2 = T_\mathrm{out}^1 /2$
\item batch normalization layer
\item layer 3: $m_{\mathrm{in} }^3=20$, $m_{\mathrm{out} }^3 = 15$, kernel size $k=4$, stride $s=2$, $T_\mathrm{out}^3 = T_\mathrm{out}^2/2$
\item batch normalization layer 
\item layer 4: $m_{\mathrm{in} }^4=15$, $m_{\mathrm{out} }^4 =10$, kernel size $k=4$, stride $s=2$, $T_\mathrm{out} ^4= T_\mathrm{out}^3/2$
\item batch normalization layer
\item layer 5: $m_{\mathrm{in} }^5=10$, $m_{\mathrm{out} }^5 = 20$, kernel size $k=4$, stride $s=2$, $T_\mathrm{out} ^5= T_\mathrm{out}^4 /2$
\item batch normalization layer
\item layer 6: $m_{\mathrm{in} }^6=20$, $m_{\mathrm{out} }^6 = 20$, kernel size $k=2$, stride $s=2$, $T_\mathrm{out} ^6= T_\mathrm{out}^5/2$
\end{itemize}
The batch normalization layer standardizes the inputs to a layer for each mini-batch to stabilize the learning. The properties of the CNN are:
\begin{itemize}
\item receptive field: 125
\item jump between two consecutive starts: 64
\item overlap: 61 at each extremity
\end{itemize}
The length of the output time series is $T/32$, with~$T=1984$.

\section{Architecture of the neural networks used for TD3 algorithm}
\label{sec:app_B}

The actor network $\mathcal{P}(\cdot,\omega)$ had the following architecture:
\begin{itemize}
\item layer 1: Linear(2, 128) + ReLU activation function
\item layer 2: Linear(128, 256) + ReLU activation function
\item layer 3: Linear(256, 2) + Tanh activation function
\end{itemize}
The output of the network was then multiplied by a user specified maximum value $c_{\rm max}$. In our experiments we set $c_{\rm max} = 10$. The critic network $\mathcal{Q}(\cdot,\cdot,\theta)$ had the following architecture:
\begin{itemize}
\item Layer 1: Linear(4, 256) + ReLU activation function
\item Layer 2: Linear(256, 256) + ReLU activation function
\item Layer 3: Linear(256, 1)
\end{itemize}

\section{Parameters for the TD3 algorithm}
\label{sec:app_C}

The TD3 algorithm used in this work is the one described in~\cite{TD3}, available at the following link: \href{https://github.com/sfujim/TD3/blob/master/TD3.py}{https://github.com/sfujim/TD3/blob/master/TD3.py}.
We have used the following parameters for the code:
\begin{itemize}
    \item $\tau = 5 \times 10^{-2}$
    \item discount: $0.99$
    \item policy\_noise: $0.2$
    \item policy\_frequency: $60$
    \item max\_action: $10$
    \item noise\_clip: $0.5$
    \item action\_dim: $2$
    \item state\_dim: $2$ 
    \item learning rate: $3 \times 10^{-4}$
\end{itemize}
The parameters for the training (game related) are the following:
\begin{itemize}
    \item dataset maximum size: $30,000$
    \item number of games: $50,000$
    \item batch size: $512$
    \item number of rounds per game $T= 600$
    \item train periodicity (in rounds per game): $d=100$
    \item exploration noise: $0.1$
    \item $\dt = 5 \times 10^{-3}$
    \item Probability of random step decay coefficient: $0.99$
    \item Random step decay period (in rounds per game): $2000$
\end{itemize}
The optimization routine used was the Adam optimization algorithm \cite{kingma2014adam} from the PyTorch library.

\bibliographystyle{plain}
\bibliography{biblio}

\end{document}